\documentclass[lettersize,journal]{IEEEtran}
\usepackage{amsmath,amsfonts}
\usepackage{algorithmic}
\usepackage{algorithm}
\usepackage{array}
\usepackage[caption=false,font=normalsize,labelfont=sf,textfont=sf]{subfig}
\usepackage{textcomp}
\usepackage{stfloats}
\usepackage{url}
\usepackage{verbatim}
\usepackage{graphicx}
\usepackage{cite}

\usepackage{multirow}
\usepackage{booktabs}

\newtheorem{theorem}{Theorem}
\newtheorem{proposition}[theorem]{Proposition}

\newtheorem{remark}{Remark}

\begin{document}

\title{Adaptive Fuzzy C-Means with Graph Embedding}


\author{Qiang~Chen, Weizhong~Yu, Feiping~Nie, and Xuelong Li

\thanks{Feiping Nie is the corresponding author.}
\thanks{Qiang Chen is with the School of Computer Science, and School of Artificial Intelligence, Optics and Electronics (iOPEN), Northwestern Polytechnical University, Xi’an 710072, Shaanxi, P. R. China. (e-mail: chenq@mail.nwpu.edu.cn).}
\thanks{Feiping Nie is with the School of Artificial Intelligence, OPtics and ElectroNics (iOPEN), School of Computer Science, Northwestern Polytechnical University, Xi’an 710072, P.R. China, and also with the Key Laboratory of Intelligent Interaction and Applications (Northwestern Polytechnical University), Ministry of Industry and Information Technology, Xi’an 710072, P.R. China (email: feipingnie@gmail.com).}
\thanks{Weizhong Yu is with the School of Artificial Intelligence, OPtics and ElectroNics (iOPEN), Northwestern Polytechnical University, Xi’an 710072, P.R. China, and also with the Key Laboratory of Intelligent Interaction and Applications (Northwestern Polytechnical University), Ministry of Industry and Information Technology, Xi’an 710072, P.R. China (email: yuwz05@mail.xjtu.edu.cn).}
\thanks{Xuelong Li is with the Institute of Artificial Intelligence (TeleAI), China Telecom Corp Ltd, 31 Jinrong Street, Beijing 100033, P. R. China (email: li@nwpu.edu.cn).}

}


\markboth{}
{Shell \MakeLowercase{\textit{et al.}}: A Sample Article Using IEEEtran.cls for IEEE Journals}


\maketitle

\begin{abstract}
Fuzzy clustering algorithms can be roughly categorized into two main groups: Fuzzy C-Means (FCM) based methods and mixture model based methods. However, for almost all existing FCM based methods, how to automatically selecting proper membership degree hyper-parameter values remains a challenging and unsolved problem. Mixture model based methods, while circumventing the difficulty of manually adjusting membership degree hyper-parameters inherent in FCM based methods, often have a preference for specific distributions, such as the Gaussian distribution. In this paper, we propose a novel FCM based clustering model that is capable of automatically learning an appropriate membership degree hyper-parameter value and handling data with non-Gaussian clusters. Moreover, by removing the graph embedding regularization, the proposed FCM model can degenerate into the simplified generalized Gaussian mixture model. Therefore, the proposed FCM model can be also seen as the generalized Gaussian mixture model with graph embedding. Extensive experiments are conducted on both synthetic and real-world datasets to demonstrate the effectiveness of the proposed model.
\end{abstract}


\section{Introduction}
Fuzzy clustering algorithms have been extensively utilized to uncover potential latent structures within data. Typically, fuzzy clustering algorithms can be roughly divided into two main categories: FCM based methods and mixture model based methods.

FCM stands as one of the most classic fuzzy clustering models, forming the cornerstone for numerous existing fuzzy clustering algorithms. The fundamental concept behind FCM is to regulate the fuzziness of membership degrees through hyper-parameters such as weighting exponent \cite{bezdek2013pattern, xuefcm}, entropy regularization \cite{li1995maximum}, or quadratic term \cite{miyamoto1998fuzzy} hyper-parameters. Although various FCM variants have been developed \cite{xu2016robust, ruspini2019fuzzy}, the challenge still persists in finding a reliable method for automatically selecting the optimal values of these membership degree hyper-parameters. In almost all existing FCM based clustering algorithms, the membership degree hyper-parameters are adjusted through historical experience or experimental methods. Unfortunately, these approaches are often inefficient and may lead to poor clustering results.

Another drawback of FCM is that it can only achieve good performance on data with spherical clusters, whereas real-world cluster shapes are often complex and non-spherical, such as ellipsoidal and non-Gaussian. To improve the performance of FCM on data with ellipsoidal clusters, Mahalanobis distance was introduced into FCM \cite{zhao2015mahalanobis, chen2023rooted}. For non-Gaussian clusters, some kernel-based FCM algorithms have been developed \cite{chen2011multiple, lu2020multi, zeng2020kernelized}. These algorithms can project data from the low-dimensional space to a high-dimensional space, facilitating linear classification of non-Gaussian clusters. However, kernel-based methods are susceptible to noises, and selecting a proper kernel is also a challenging problem.

Recently, graph embedding has become a frequently used technique in machine learning fields because it can help algorithms handle non-Gaussian data \cite{xia2022tensorized, xia2023graph, yang2022multiview, zhao2021graph}. Spectral clustering is a classic graph based clustering algorithm. It involves constructing graphs to represent samples using an affinity matrix, followed by the application of the K-means algorithm to these graph representations of the data \cite{macqueen1967some, ng2001spectral, von2007tutorial}. Compared to kernel embedding methods, graph embedding methods can better capture the local structure of data and may yield more satisfactory results. Inspired by spectral clustering, lots of graph based clustering algorithms were developed. In order to obtain the optimal graph similarity matrix, Nie et al. proposed a novel model that can automatically learn proper weights of the similarity matrix \cite{nie2014clustering}. To generate a graph with clusters, Han et al. proposed a spectral clustering model that incorporates orthogonal and nonnegative constraints \cite{han2017orthogonal}. Moreover, their approach enables the direct acquisition of final cluster labels, eliminating the necessity for post-processing. Pei et al. revisited the unified framework of K-means and ratio-cut spectral clustering, and proposed an efficient clustering algorithm based on the framework \cite{pei2020efficient}. Recently, a novel clustering algorithm utilizing bipartite graphs was also developed \cite{nie2023fast}, which posses an excellent clustering performance but comparatively low computational complexity.

Owing to the excellent performance of graph based methods on non-Gaussian data, various graph based fuzzy clustering algorithms have recently been developed as well. Locality preservation \cite{he2003locality} is a proficient graph based dimensional reduction method, and Zhou et al. introduced this method into FCM to enhance clustering performance \cite{zhou2021projected}. To better handel data with balanced clusters, Liu et al. proposed a balance regularization to suppress unbalanced classes. Recently anchor graph is getting a lot of attention, and some anchor graph based clustering methods have been developed \cite{10183380, 9436033}. Wang et al. proposed to constraint the spareness of membership degree matrix utilizing $L_0$-norm \cite{wang2022projected}, which may suppress the influence of outliers on graph weights learning and improve clustering performance.
Besides, Zhao et al. proposed a novel FCM model featured on shrunk pattern based manifold learning \cite{zhao2021robust}, which can be also seen as a graph based method with the same graph weights.

Mixture model based methods can be viewed as the other branch of fuzzy clustering. A mixture model can be seen as a linear combination of multiple probability distributions and is usually optimized by the EM algorithm \cite{mclachlan2019finite}. The Gaussian mixture model is one of the most popular model based methods \cite{reynolds2009gaussian}. However, the Gaussian mixture model assumes that data is drawn from the Euclidean space. In reality, naturally occurring data may reside on or close to an underlying sub-manifold. For considering the sub-manifold structure, Laplacian regularized Gaussian mixture model and locally consistent Gaussian mixture model were developed \cite{Liu10LCGMM, He11LapGMM}. Generally, heavy-tailed distributions are more robust than the Gaussian distribution, such as the generalized Gaussian distribution \cite{pascal2013parameter} and the Student's $t$-distribution \cite{ahsanullah2014normal}. Therefore, some heavy-tailed distributions based mixture models were proposed to improve clustering performance on polluted datasets \cite{zhang2010robust, dang2015mixtures}.

By summarizing the above analysis we realize that, for almost all FCM based clustering methods, the problem of automatically learning membership degree hyper-parameters is still challenging and not well-resolved. Additionally, most mixture model based methods only focus on data with specific distributions, i.e., Gaussian distribution, and lack the ability of handling non-Gaussain data. The main contributions of this paper are summarized as follows:
\begin{itemize}
\item The proposed FCM model introduces a membership hyper-parameter adaptive learning mechanism that can automatically learn a proper value for the membership degree hyper-parameter.
\item By introducing the graph embedding regularization term, the proposed FCM model can handle data with non-Gaussian clusters, which also proves the transferability of the membership hyper-parameter adaptive learning mechanism.
\item By removing the graph embedding regularization, the proposed FCM model can degenerate into a simplified generalized Gaussian mixture model. Therefore, the proposed FCM model can be also viewed as a mixture model with graph embedding.
\item An efficient alternating optimization strategy with closed solutions is provided to optimize the proposed FCM model.
\end{itemize}

{\bf Notations: } Throughout the paper, let $\mathbb{R}$, $\mathbb{R}^+$, $\mathbb{R}^{n}$ and $\mathbb{R}^{n\times m}$ denote the sets of real numbers, positive real numbers, length-$n$ vectors and size $n\times m$ matrices, respectively. Suppose data matrix $X=[x_1, x_2, ..., x_n]\in \mathbb{R}^{d\times n}$ consists of $n$ samples with $c$ clusters, and each sample $x_i$ has $d$ features. For matrix $A$, the element in the $i$-th row and the $j$-th column of $A$ is denoted by $a_{ij}$, and trace of $A$ is denoted by $Tr(A)$.
The adjacency matrix of an undirected weighted graph is defined as $W$ and the degree matrix is defined as $D$. Then the Laplacian matrix is defined as $L=D-W$, and the normalized one is defined as $\hat{L}=D^{-1/2}LD^{-1/2}$. The $L_2$-norm of vector $v$ is denoted by $||v||_2$. An identity matrix with $n$ diagonal elements is denoted by $I_n$.
\section{Related Works}
In this section, we introduce some important related works, including FCM, spectral clustering and the generalized Gaussian mixture model.
\subsection{FCM}
FCM is an extremely significant fuzzy clustering model as it forms the basis for plenty of fuzzy clustering algorithms. According to the methods of controlling fuzziness, FCM can be roughly divided into three types: weighting exponent based, entropy regularization based, and quadratic term based. The objective function of FCM with entropy regularization is shown below.
\begin{equation}\label{e1}
\begin{gathered}
\min_{U,V} \sum_{i=1}^{n}\sum_{j=1}^{c} u_{ij} ||x_i-v_j||_2^2 + \frac{1}{\gamma} u_{ij}\log u_{ij}, \\ {\rm s.t.} \ \ \sum_{j=1}^{c}u_{ij}=1, \ \ 0< u_{ij}< 1
\end{gathered}
\end{equation}
where $U\in\mathbb{R}^{n\times c}$ denotes the membership degree matrix, $V=\{v_1,v_2,...,v_c\}$ denotes the set of cluster centers, $\gamma\in \mathbb{R}^+$ denotes the entropy regularization hyper-parameter, and $n$ and $c$ denote the numbers of samples and clusters, respectively.
\begin{remark}
In almost all existing FCM with entropy regularization algorithms, the entropy regularization hyper-parameter $\gamma$ can not be automatically updated. Typically, the value of $\gamma$ is adjusted by historical experience or experimental results.
\end{remark}
\subsection{Spectral Clustering}
Spectral clustering is a popular clustering algorithm for its effectiveness on handling non-Gaussian clusters. The fundamental concept behind spectral clustering is the similarity graph that encapsulates pairwise similarities among data points, with higher similarity weights indicating closer proximity. By strategically eliminating weaker edges, the algorithm extracts c independent sub-graphs, ultimately yielding c clusters. However, the optimization of spectral clustering is NP-hard, and it usually resorts to optimizing the following relaxed problems.
\begin{equation}\label{spectral}
\begin{gathered}
\min_{F} Tr(F^TLF), \ \ {\rm s.t. } \ \ F^TF=I_c
\end{gathered}
\end{equation}
where $F\in\mathbb{R}^{n\times c}$ is the indicator matrix with $c$ clusters and $L\in\mathbb{R}^{n\times n}$ is the Laplacian matrix. The objective function in Eq. (\ref{spectral}) can be optimized by computing the corresponding eigenvectors of the $c$ minimum eigenvalues. Subsequently, performing K-means on $F$ allows us to obtain $c$ clusters. According to the property of Laplacian matrix, the objective function in Eq. (\ref{spectral}) can be reformulated as
\begin{equation}\label{spectral2}
\begin{gathered}
\min_{F} \frac{1}{2} \sum_{i=1}^{n}\sum_{j=1}^{n} w_{ij}||f_i-f_j||_2^2  \ \ {\rm s.t. } \ \ F^TF=I_c
\end{gathered}
\end{equation}
where $w_{ij}$ is the element of the adjacency matrix $W$ representing the similarity between sample $x_i$ and sample $x_j$, and $f_i\in\mathbb{R}^{1\times c}$ is the $i$-th row of the $F$. Therefore, the objective function in Eq. (\ref{spectral2}) can also be seen as a manifold learning algorithm, which projects sample $x_i$ form the original manifold as $f_i$ into the new manifold while preserving  the similarities among samples. If the Laplacian matrix $L$ is replaced with the normalized Laplacian matrix $\hat{L}$, the objective function of R-cut spectral clustering in Eq. (\ref{spectral}) is transformed into that of N-cut spectral clustering.

\subsection{Generalized Gaussian Mixture Model}
The Kotz-type distribution is a member of the elliptically contoured distribution family, and the generalized Gaussian distribution is a special case of the Kotz-type distribution \cite{kotz1975multivariate, fang2018symmetric, nadarajah2003kotz}.
Let $x\in \mathbb{R}^d$ be a $d$-dimensional random vector, and the probability density function of the generalized Gaussian distribution is defined as follows.
\begin{equation}\label{e4}
\begin{gathered}
g(x|\theta )\! =\!  \frac{{\beta \Gamma (\frac{d}{2}){m^{\frac{d}{{2\beta }}}}}}{{{\pi ^{\frac{d}{2}}}\Gamma (\frac{d}{{2\beta }})}}  |\Sigma {|^{ - \frac{1}{2}}} \times  \quad\qquad\qquad  \\
 \qquad\qquad \exp\! \left\{\!  - m  {{\left[ {{{(x\! - \!v)}^T}{\Sigma ^{ - 1}}(x\! - \!v)} \right]}^\beta } \! \right\} \\
\end{gathered}
\end{equation}
where $\theta = \{v,\Sigma,\beta,m\}$ denotes the set of parameters, $v\in\mathbb{R}^d$ denotes the mean, $\Sigma\in\mathbb{R}^{d\times d}$ denotes the positive definite covariance matrix, $\beta\in\mathbb{R}^+$ denotes the shape parameter, and $m\in \mathbb{R}^+$ denotes the scale parameter.

The generalized Gaussian mixture model can be seen as a linear combination of multiple generalized Gaussian components, and the probability distribution function can be written as
\begin{equation}\label{d}
\begin{gathered}
f(x|\Theta ) = \mathop \sum \limits_{j = 1}^c {\alpha _j}g(x|{\theta _j}), \ \ {\rm s.t.}\sum_{j=1}^{c}\alpha_j=1
\end{gathered}
\end{equation}
where $\Theta  = \{ \alpha ,V,\Sigma ,\beta ,m\} $ denotes the set of all parameters in the MGGD mixture model, ${\theta _j} = \left\{ {{v_j},{\Sigma _j},\beta ,m} \right\}$ denotes the parameter set of the $j$-th component, $\alpha  = \{ {\alpha _1},{\alpha _2},..,{\alpha _c}\}$ denotes the set of mixing coefficients, and $c$ denotes the number of components.

For estimating the parameters of a probability distribution, the maximum likelihood estimation method is frequently used, and the log-likelihood function is constructed as follows.
\begin{equation}\label{mle}
\begin{gathered}
  L(\Theta |X) \! \! =\! \! \mathop \sum \limits_{i = 1}^n \!  \log f({x_i}|\Theta ) \! = \!\! \mathop \sum \limits_{i = 1}^n \!\log\! \mathop \sum \limits_{j = 1}^c \! {\alpha _j}g({x_i}|{\theta _j}), \\
   {\rm s.t.}\sum_{j=1}^{c}\alpha_j=1
\end{gathered}
\end{equation}

However, optimizing the above objective function presents challenges due to the presence of sum in the logarithm. The Expectation-Maximization (EM) algorithm \cite{bilmes1998gentle}, a probability-based optimization method, is a frequently used method to addresses this difficulty. Instead of directly optimizing the original objective function, EM introduces a latent variable. The optimization process then shifts towards maximizing the Q function, a derived auxiliary function.

\section{The Proposed Method}
\subsection{Formulation}
In this subsection, we initially introduce how the equivalent connection between FCM and the generalized Gaussian mixture model is constructed. Then, based on this, we show how the problem of automatically learning membership degree hyper-parameters in FCM is solved. Furthermore, we introduce how the graph is embedded into FCM to help deal with non-Gaussian clusters, and finally present the objective function of the proposed model.

{\bf Equivalence and hyper-parameter learning:}
Since the log-likelihood function of the generalized Gaussian mixture model involves a sum inside the logarithm, optimizing this objective function requires using the EM algorithm. Notably, the EM algorithm, being grounded in probability theory, diverges significantly from the optimization strategy employed in FCM. In order to build the connection between FCM and the generalized Gaussian mixture model, the equivalent objective function of the generalized Gaussian mixture model is constructed as follows.
\begin{proposition}\label{pro1}
The update equations of the maximum log-likelihood function in Eq. (\ref{mle}) for the generalized Gaussian mixture model, which is optimized through the EM algorithm,  are equivalent to the update equations of the following objective function.
\begin{equation}\label{pro1_e1}
\begin{gathered}
\hspace{-0.1cm}
\mathop {\min }\limits_{U,\alpha, V, \Sigma, \beta ,m} \mathop \sum \limits_{i = 1}^n \mathop \sum \limits_{j = 1}^c {u_{ij}}\bigg\{ m{{ {{\left[ {{{(x_i  - {v_j})}^T}{\Sigma_j^{ - 1}}(x_i - {v_j})} \right]}} }^\beta } \ \
\\ \quad + \frac{1}{2}\log |{\Sigma _j}| - \log {\alpha _j} - \log \frac{{\beta \Gamma (\frac{d}{2}){m^{\frac{d}{{2\beta }}}}}}{{{\pi ^{\frac{d}{2}}}\Gamma (\frac{d}{{2\beta }})}} + \log {u_{ij}} \bigg\}, \\
\ \ {\rm s.t.} \ \ \mathop \sum \limits_{j = 1}^c {\alpha _j} = 1, \ \ 0 < {\alpha _j} < 1, \\
       \ \ \ \ \ \ \ \    \mathop \sum \limits_{j = 1}^c {u_{ij}} = 1, \ \ 0 < {u_{ij}} < 1
\end{gathered}
\end{equation}
where $u_{ij}$ is the membership degree denoting the probability of sample $x_i$ being assigned to the $j$-th cluster, $\alpha  = \{ {\alpha _1},{\alpha _2},..,{\alpha _c}\}$, $V = \{ {v_1},{v_2},..,{v_c}\} $ and $\Sigma  = \{ {\Sigma _1},{\Sigma _2},..,{\Sigma _c}\} $ denote the sets of mixing coefficients, means and scale matrices, and $\beta$ and $m$ denote the shape parameter and the scale parameter in the generalized Gaussian mixture model. To concentrate on the main idea, {\bf the proof for Proposition \ref{pro1} is moved to the Appendix.}
\end{proposition}

If we consider only the membership degree matrix $U$ and the set of means $V$ as variables in Eq. (\ref{pro1_e1}), treating other variables as constants, i.e., $\Sigma_j=I_d$, $\alpha_j=1/c$, $\beta=1$ and $m\in\mathbb{R}^+$, we obtain the objective function of the simplified generalized Gaussian mixture model as follows.
\begin{equation}\label{sm}
\begin{gathered}
\min_{U,V} \sum_{i=1}^{n}\sum_{j=1}^{c} u_{ij} m ||x_i-v_j||_2^2 + u_{ij}\log u_{ij}, \\ {\rm s.t.} \ \ \sum_{j=1}^{c}u_{ij}=1, \ \ 0< u_{ij}< 1
\end{gathered}
\end{equation}
Then let us look back at the objective function of FCM in Eq. (\ref{e1}). Since the membership degree hyper-parameter $\gamma$ in FCM is a positive constant, the objective function of FCM in Eq. (\ref{e1}) can be transformed by multiplying $\gamma$ as follows.
\begin{equation}\label{fcm1}
\begin{gathered}
\min_{U,V} \sum_{i=1}^{n}\sum_{j=1}^{c} u_{ij} \gamma ||x_i-v_j||_2^2 + u_{ij}\log u_{ij}, \\ {\rm s.t.} \ \ \sum_{j=1}^{c}u_{ij}=1, \ \ 0< u_{ij}< 1
\end{gathered}
\end{equation}
By comparing the objective function of the simplified generalized Gaussian mixture model in Eq. (\ref{sm}) with that of FCM in Eq. (\ref{fcm1}), we can observe that the membership degree hyper-parameter $\gamma$ in FCM serves the same purpose as the scale parameter $m$ in the generalized Gaussian mixture model. However, the scale parameter $m$ in the generalized Gaussian model can be learned automatically. Therefore, it is possible to learn the membership degree hyper-parameter $\gamma$ in FCM in the same manner.

In Eq. (\ref{pro1_e1}), if we replace $m$ with $\gamma$, consider $U$, $V$ and $\gamma$ as variables, and set $\Sigma_j=I_d$, $\alpha_j=1/c$ and $\beta=1$, we obtain an FCM objective function that can automatically learn the membership degree hyper-parameter $\gamma$ as follows.
\begin{equation}\label{afcm}
\begin{gathered}
\mathop {\min } \limits_{U,V,\gamma} \! \sum\limits_{i = 1}^n \!  \sum\limits_{j = 1}^c \!  u_{ij} \gamma||x_i - {v_j}||_2^2  + u_{ij} \log {u_{ij}}  - n\log {\gamma^{\frac{d}{2}}}, \\
{\rm{              }}{\rm s.t.} \ {\rm{  }}\mathop \sum \limits_{j = 1}^c {u_{ij}} = 1, \ {\rm{  }}0 < {u_{ij}} < 1
\end{gathered}
\end{equation}
where $d$ is the number of features for sample $x_i$. Note that, the FCM objective function in Eq. (\ref{afcm}) can be also seen as a simplified generalized Gaussian mixture model, because they have the same update equations.


\begin{figure}[!t]
\centering
\subfloat[]{\includegraphics[width=1.2in]{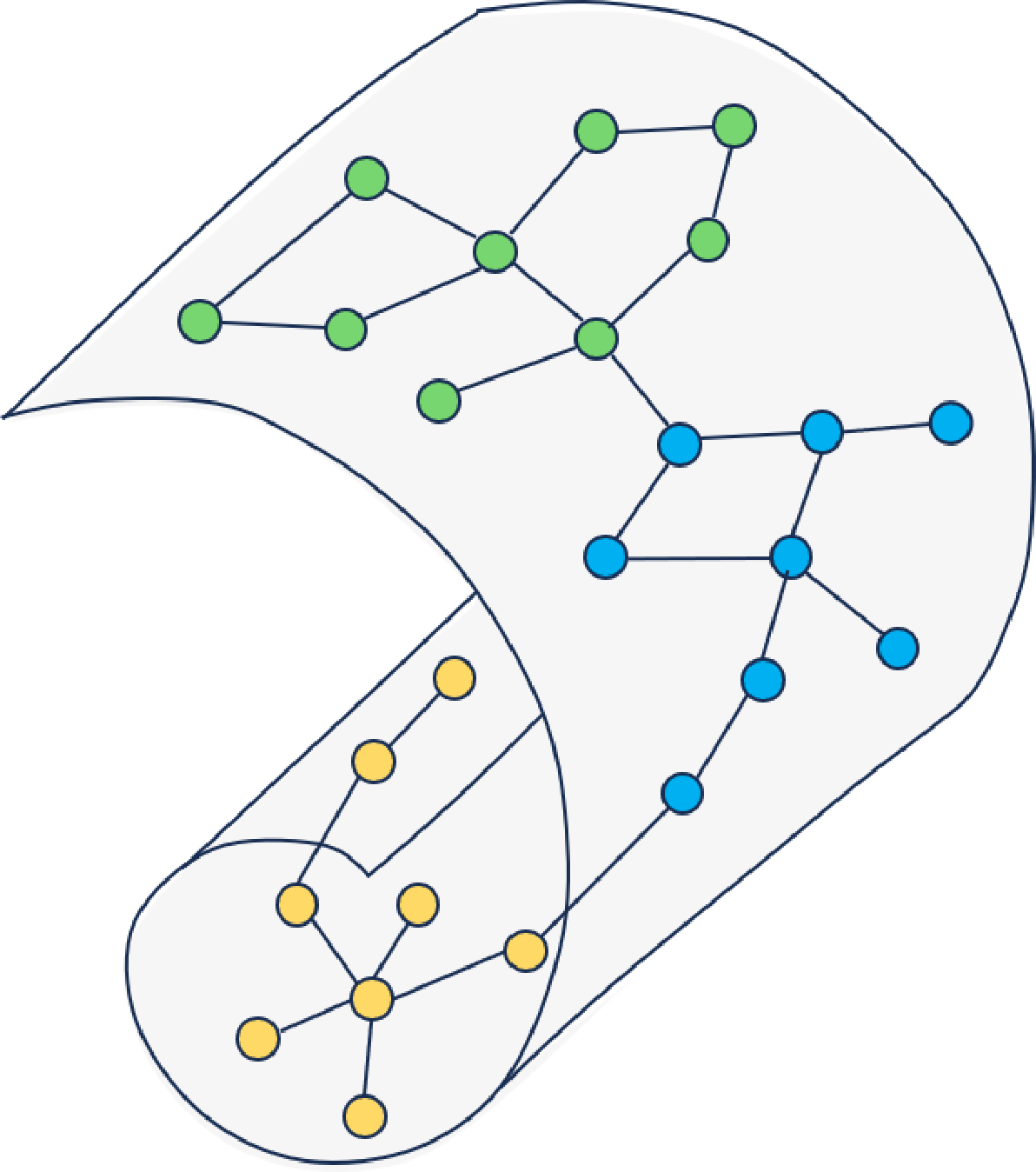}%
\label{fig_first_case}}
\hfil
\subfloat[]{\includegraphics[width=0.9in]{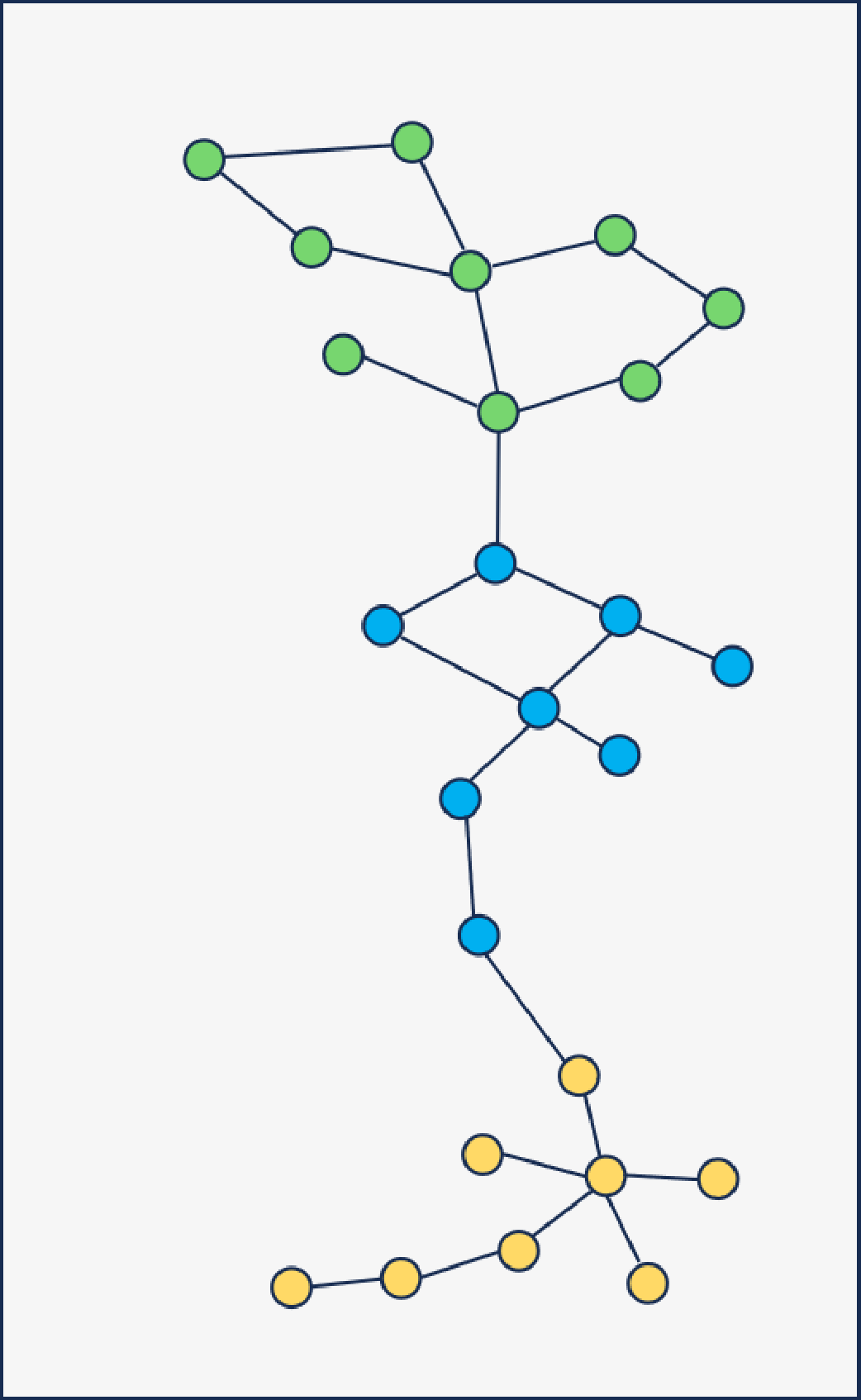}%
\label{fig_second_case}}
\caption{Illustration on graph based manifold learning. For data residing in a complex manifold, as depicted in (a), Euclidean distance may not accurately capture the true relationships between samples. However, by projecting the data according to the similarity graph into a simpler manifold, as illustrated in (b), the newly learned manifold facilitates easier clustering compared to the original manifold.}
\label{fig1}
\end{figure}

{\bf Graph embedding and the proposed model:} A drawback of FCM is that it may not perform well with non-spherical clusters because it relies directly on Euclidean distance. However, real-world data are often non-spherical and non-Gaussian. A frequently used method for handling non-Gaussian data is graph based manifold learning. It constructs a similarity graph of the data and then projects them according to these similarities from the original manifold into a new manifold where the data can be more effectively clustered. An illustration of graph based manifold learning is shown in Fig. \ref{fig1}.

A key point for graph based manifold learning is graph construction. Membership degree $u_{ij}$ in FCM denotes the probability of the $i$-th sample belonging to the $j$-th cluster, but it can also be used to reveal the latent relationships between samples. Therefore, it is possible to construct a new similarity graph, based on the membership degrees, to help complete the original similarity graph. If the update equation of cluster center $v_j$ can be represented as $v_j={\sum_{i=1}^{n}u_{ij}x_i}/{\sum_{i=1}^{n}u_{ij}}$, we have the following equation.
\begin{equation}\label{emm}
\begin{gathered}
\sum_{i=1}^n\sum_{j=1}^cu_{ij}||x_i-v_j||_2^2 \mathop  = Tr\left[ X(I_n-UBU^T)X^T \right]
\end{gathered}
\end{equation}
where $X=[x_1,x_2,...,x_n]\in\mathbb{R}^{d\times n}$ is the data matrix, $U\in\mathbb{R}^{n\times c}$ is the membership degree matrix, and $B\in\mathbb{R}^{c\times c}$ is a diagonal matrix with the $k$-th diagonal element being $b_{kk}= 1/\sum_{i=1}^{n}u_{ik}$. To concentrate on the main idea, {\bf the proof for Eq. (\ref{emm}) is moved to the Appendix.}

According to the graph theory proposed by Liu et al. \cite{liu2010large}, we know that the matrix $I_n-UBU^T$ in Eq. (\ref{emm}) can be a Laplacian matrix on the anchor graph, and the corresponding adjacency matrix is $\{\hat{W}|\hat{w}_{ij} = \sum_{k=1}^{c} u_{ik}u_{jk} / b_{kk} \}$. Then we can conclude that
\begin{equation}\label{emm1}
\begin{gathered}
Tr\left[X(I_n-UBU^T)X^T\right] = \frac{1}{2} \sum_{i=1}^n\sum_{j=1}^n \hat{w}_{ij}||x_i-x_j||_2^2
\end{gathered}
\end{equation}
where $\hat{w}_{ij}$ reflects the relationship between $x_i$ and $x_j$. Therefore, if we consider data matrix $X$ as a variable and denote it by $\tilde{X}$, it is possible to project the data into a new manifold by optimizing the following objective function.
\begin{equation}\label{emm2}
\begin{gathered}
\min_{\tilde{X}} Tr\left[\tilde{X}(I_n-UBU^T)\tilde{X}^T\right]
\end{gathered}
\end{equation}
where $\tilde{X}\in \mathbb{R}^{ \tilde{d}\times n}$ denotes the newly learned data matrix, and $\tilde{d}$ denotes the dimension of the newly learned sample $\tilde{x}_i$. Utilizing the objective function in Eq. (\ref{emm2}) can make the projected data more compact within the same cluster, but sometimes the learned Laplacian matrix $I_n-UBU^T$ may be inaccurate and lead to incorrect projection. Therefore, we attempt to introduce the normalized Laplacian matrix $\hat{L}$ (see the definition in Notations) into the objective function in Eq. (\ref{emm2}) to help obtain a better manifold. The specific formulation is given as follows.
\begin{equation}\label{emm3}
\begin{gathered}
\min_{\tilde{X}}= Tr\left[\tilde{X}(I_n-UBU^T)\tilde{X}^T\right] + \lambda Tr(\tilde{X}\hat{L}\tilde{X}^T)
\end{gathered}
\end{equation}
where $\lambda\in\mathbb{R}^+$ is used to balance the influence of the second term. Then, {\bf the objective function of the proposed AFCM (Adaptive Fuzzy C-Means with graph embedding) model is shown as follows. }
\begin{equation}\label{obj}
\begin{gathered}
\mathop {\min } \limits_{U,V,\gamma,\tilde{X}} \! \sum\limits_{i = 1}^n \!  \sum\limits_{j = 1}^c \!  u_{ij} \gamma||\tilde{x}_i - {v_j}||_2^2 + \lambda Tr(\tilde{X}\hat{L}\tilde{X}^T) \\
\qquad\quad \ \  + u_{ij} \log {u_{ij}}  - n\log {\gamma^{\frac{\tilde{d}}{2}}}, \\
{\rm{              }}{\rm s.t.} \ {\rm{  }} \tilde{X}\tilde{X}^T = I_{\tilde{d}} \, , \mathop \sum \limits_{j = 1}^c {u_{ij}} = 1, \ {\rm{  }}0 < {u_{ij}} < 1
\end{gathered}
\end{equation}
where $\tilde{d}$ is the dimension of the projected sample $\tilde{x_i}$, and in this paper we directly set $\tilde{x_i}=c$, and the orthogonal constraint $\tilde{X}\tilde{X}^T = I_{\tilde{d}}$ is introduced to avoid trivial solutions. Note that the model in Eq. (\ref{afcm}) can be viewed as a degenerate form of the proposed model in Eq. (\ref{obj}).

\subsection{Optimization}
There are four variables to be updated for the proposed objective function in Eq. (\ref{obj}).

When we update $U$ while keeping other variables fixed, the objective function in Eq. (\ref{obj}) can be reformulated as

\begin{equation}\label{op1}
\begin{gathered}
	\min_U \sum_{i=1}^n{\sum_{j=1}^c{u_{ij} \gamma ||(\tilde{x}_i-v_j)||_{2}^{2}+u_{ij}\log \left( u_{ij} \right)}}\\
	{\rm s.t.}\mathrm{  }\sum_{j=1}^c{u_{ij}}=1,\mathrm{  }0<u_{ij}<1\\
\end{gathered}
\end{equation}
An usual way to solve problem in Eq. (\ref{op1}) is using the Lagrange multiplier method \cite{boyd2004convex}. The Lagrange function is given as
\begin{equation}\label{op2}
\begin{gathered}
L\left( U,\eta \right) =\sum_{i=1}^n\sum_{j=1}^c u_{ij}\gamma||(\tilde{x}_i-v_j)||_{2}^{2} \qquad \ \ \\
\qquad +u_{ij}\log \left( u_{ij} \right) +\sum_{i=1}^n{\eta _i ( \sum_{j=1}^c{u_{ij}}-1 )}
\end{gathered}
\end{equation}
By setting the derivative of the Lagrange function to zero with respective to $u_{ij}$, and combining it with the constraint $\sum_{j=1}^{c}u_{ij}=1$, we obtain the update solution of $u_{ij}$ as follows.
\begin{equation}\label{op3}
u_{ij}=\frac{\exp \left\{ -\gamma ||(\tilde{x}_i-v_j)||_{2}^{2} \right\}}{\sum_{j=1}^c{\exp \left\{ -\gamma ||(\tilde{x}_i-v_j)||_{2}^{2} \right\}}}
\end{equation}

When we update $V$ while keeping other variables fixed, the objective function in Eq. (\ref{obj}) can be reformulated as
\begin{equation}\label{op4}
\min_V \sum_{i=1}^n{\sum_{j=1}^c{u_{ij}||(\tilde{x}_i-v_j)||_{2}^{2}}}
\end{equation}
By setting the derivative of the objective function in Eq. (\ref{op4}) to zero with respect to $v_j$ we obtain the update equation of $v_j$ as follows.
\begin{equation}\label{op5}
v_j=\frac{\sum_{i=1}^n{u_{ij}\tilde{x}_i}}{\sum_{i=1}^n{u_{ij}}}
\end{equation}

When we update $\gamma$ while keeping other variables fixed, the objective function in Eq. (\ref{obj}) can be reformulated as
\begin{equation}\label{op6}
\min_{\gamma} \sum_{i=1}^n{\sum_{j=1}^c{u_{ij}\gamma ||(\tilde{x}_i-v_j)||_{2}^{2}-n\log \gamma ^{\frac{\tilde{d}}{2}}}}
\end{equation}
By setting the derivative of the objective function in Eq. (\ref{op6}) to zero with respect to $\gamma$ we obtain the update equation of $\gamma$ as follows.
\begin{equation}\label{op7}
\gamma =\,\,\frac{\frac{1}{2} \tilde{d} n}{\sum_{i=1}^n{\sum_{j=1}^c{u_{ij}||(\tilde{x}_i-v_j)||_{2}^{2}}}}
\end{equation}

When we update $\tilde{X}$ while keeping other variables fixed, the objective function in Eq. (\ref{obj}) can be reformulated as
\begin{equation}\label{op8}
\begin{aligned}
	&\min_{\tilde{X}} \sum_{i=1}^n{\sum_{j=1}^c{u_{ij}\gamma ||\tilde{x}_i-v_j||_{2}^{2}}}+\lambda Tr(\tilde{X}\hat{L}\tilde{X}^T)\\
	\Leftrightarrow \,\,&\min_{\tilde{X}} \gamma Tr\left[\tilde{X}(I_n-UBU^T)\tilde{X}^T\right]+\lambda Tr(\tilde{X}\hat{L}\tilde{X}^T)\\
	\Leftrightarrow \,\,&\min_{\tilde{X}} Tr\left\{ \tilde{X}\left[ \gamma (I_n-UBU^T)+\lambda \hat{L} \right] \tilde{X}^T \right\}\\
	&  \qquad\qquad\quad   \mathrm{s.t.} \ \  \tilde{X}\tilde{X}^T=I_{\tilde{d}}\mathrm{ }\\
\end{aligned}
\end{equation}
The proof for Eq. (\ref{op8}) is similar to that for Eq. (\ref{emm}), which has been provided in the Appendix. According to the Rayleigh quotient \cite{horn2012matrix}, the problem in Eq. (\ref{op8}) can be solved by selecting the $c$ minumum eigenvectors of matrix $ \gamma (I_n-UBU^T)+\lambda \hat{L} $.

Up to this point, the optimization process has been completed. Algorithm \ref{algorithm1} presents the pseudo code for optimizing the AFCM model defined in Eq. (\ref{obj}). Algorithm \ref{algorithm2} presents the pseudo code for optimizing the degenerate AFCM model defined in Eq. (\ref{afcm}).

\begin{algorithm}[t]
\caption{Algorithm for AFCM in Eq. (\ref{obj})}
\begin{algorithmic}[1]
\label{algorithm1}
\STATE {\bf Input:} Data matrix $X$, cluster number $c$, parameter $\lambda$, and normalized Laplacian matrix $\hat{L}$.
\STATE {\bf Initialization:} Membership degree matrix $U$, cluster center $V$, adaptive hyper-parameter $\gamma$ and projected dimension $\tilde{d} = c$.
\WHILE{not converge}
    \STATE $\tilde{X}$ \! $\leftarrow$ \! ${\rm eigenvector}$ [$ \gamma (I_n-UDU^T)+\lambda \hat{L}$, $c$ smallest vectors];
    \STATE Update $v_j$ by Eq. (\ref{op5});
    \STATE Update $\gamma$ by Eq. (\ref{op7});
    \STATE Update $u_{ij}$ by Eq. (\ref{op3});
\ENDWHILE
\STATE {\bf Output:} membership degree matrix $U$.
\end{algorithmic}
\end{algorithm}

\begin{algorithm}[t]
\caption{Algorithm for degenerated AFCM in Eq. (\ref{afcm})}
\begin{algorithmic}[1]
\label{algorithm2}
\STATE {\bf Input:} Data matrix $X$ and cluster number $c$.
\STATE {\bf Initialization:} Membership degree matrix $U$, projected data matrix $\tilde{X}=X$ and projected dimension $\tilde{d}=d$.
\WHILE{not converge}
    \STATE Update $v_j$ by Eq. (\ref{op5});
    \STATE Update $\gamma$ by Eq. (\ref{op7});
    \STATE Update $u_{ij}$ by Eq. (\ref{op3});
\ENDWHILE
\STATE {\bf Output:} membership degree matrix $U$.
\end{algorithmic}
\end{algorithm}

\begin{figure*}[!t]
\centering
\subfloat[]{\includegraphics[width=1.95in]{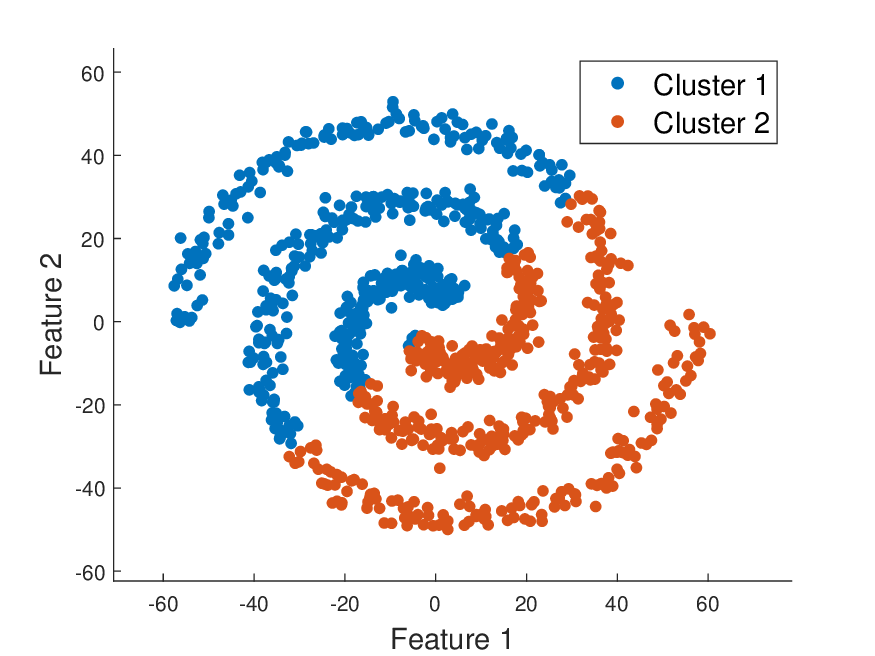}%
\hspace{-6.5mm}
\label{fig2a}}
\subfloat[]{\includegraphics[width=1.95in]{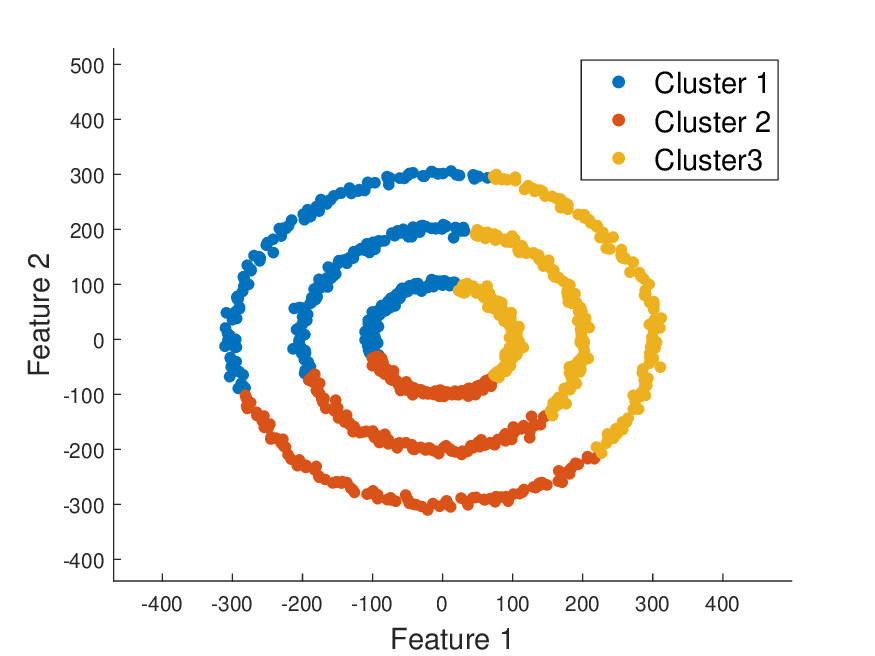}%
\hspace{-6.5mm}
\label{fig2b}}
\subfloat[]{\includegraphics[width=1.95in]{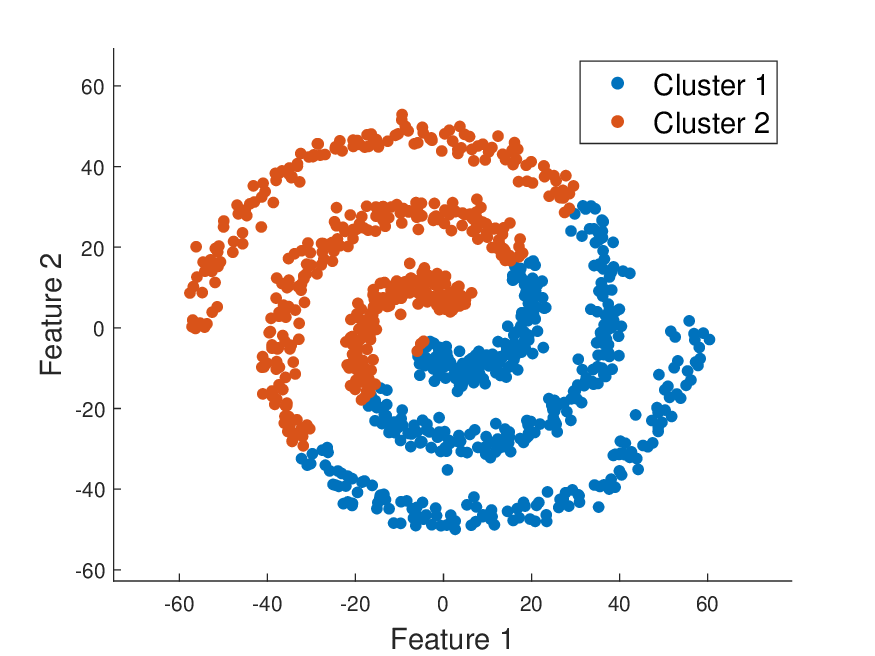}%
\hspace{-6.5mm}
\label{fig2c}}
\subfloat[]{\includegraphics[width=1.95in]{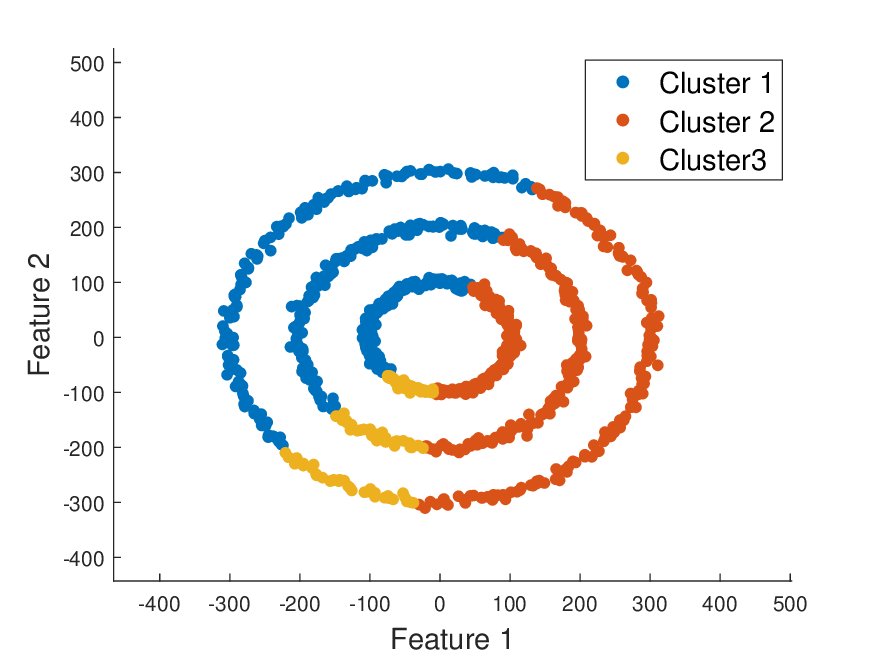}%
\label{fig2d}}
\\
\subfloat[]{\includegraphics[width=1.95in]{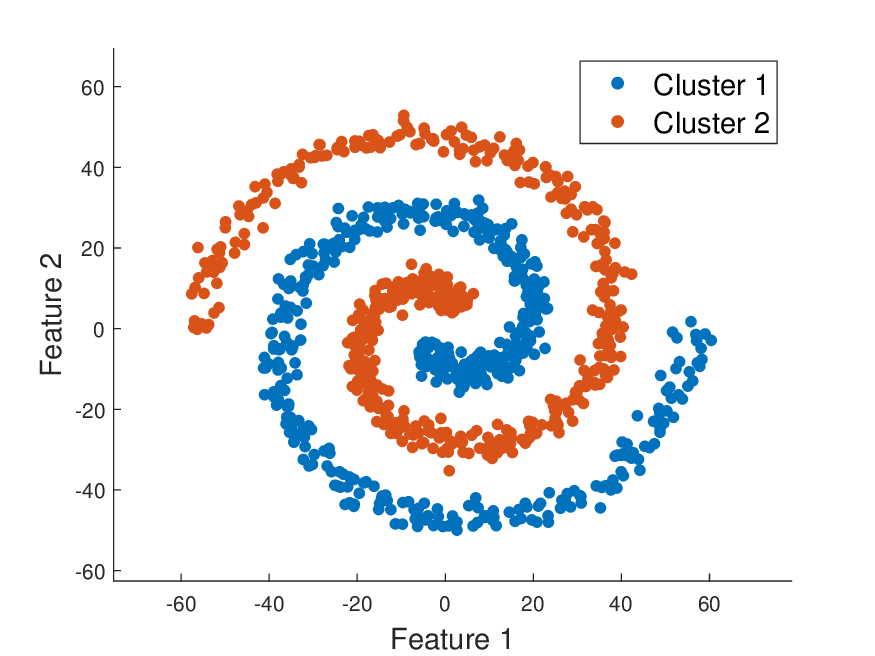}%
\hspace{-6.5mm}
\label{fig2e}}
\subfloat[]{\includegraphics[width=1.95in]{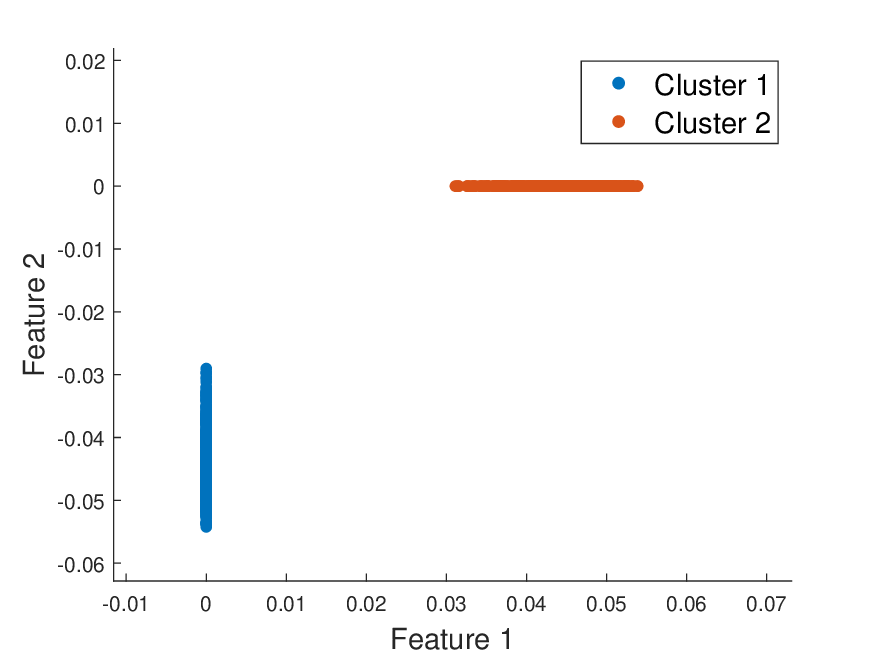}%
\hspace{-6.5mm}
\label{fig2f}}
\subfloat[]{\includegraphics[width=1.95in]{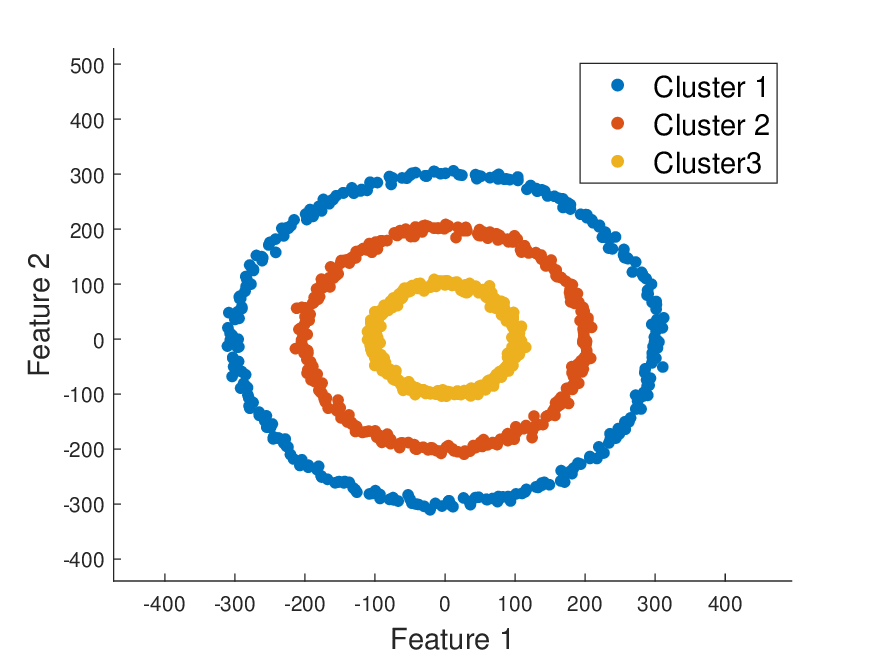}%
\hspace{-6.5mm}
\label{fig2g}}
\subfloat[]{\includegraphics[width=1.95in]{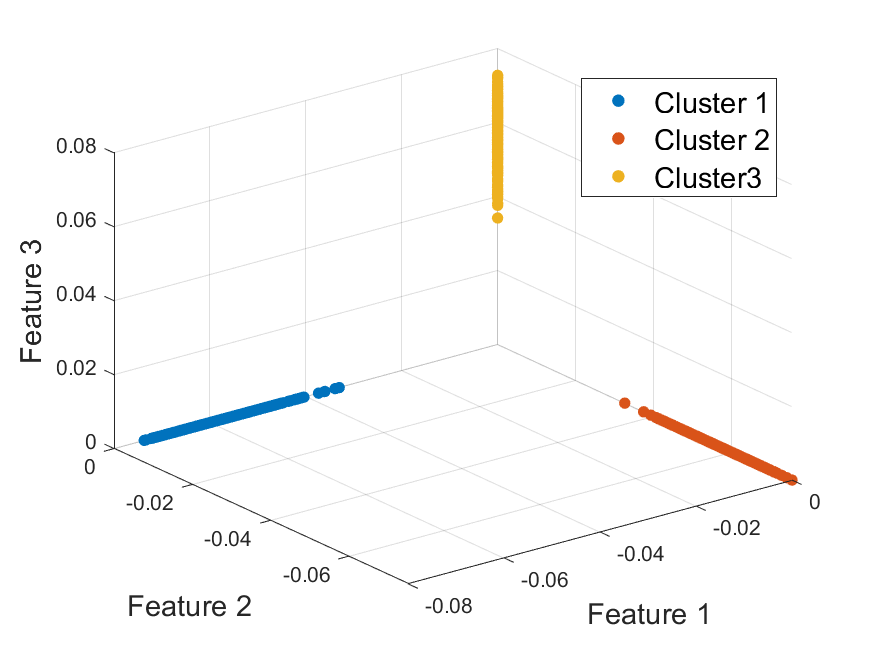}%
\label{fig2h}}
\caption{Clustering results on toy datasets. (a) and (b) depict the clustering outcomes of FCM. (c) and (d) show the clustering results of degenerate AFCM. (e)-(h) present the clustering results of AFCM, which projects data from the original manifold into a newly learned manifold and perform clustering in this new manifold rather than the original manifold.}
\label{fig_sim}
\end{figure*}

\subsection{Construction of the Affinity Matrix}
For graph embedded methods, it is important to construct a proper affinity matrix. In this paper, we use the Gaussian kernel to construct the affinity matrix as follows.
\begin{equation}
\begin{gathered}
w_{ij}=\begin{cases}\exp(-\frac{\|\boldsymbol{x}_i-\boldsymbol{x}_j\|_2^2}{2\sigma^2})&\text{if }{x}_j\in\mathcal{N}({x}_i)\\0,&\text{otherwise}\end{cases}
\end{gathered}
\end{equation}
where $\sigma$ is a predefined scalar, $\mathcal{N}({x}_i)$ denotes the set of the $k$ nearest neighbour of $x_i$. In this paper, we set the value of $\sigma=2$.

\subsection{Computational Complexity Analysis}
Let $n$ represent the number of samples, $c$ represent the number of clusters, $d$ and $\tilde{d}$ represent the original dimension and the projected dimension, and $t$ represent the number of iterative steps. For Algorithm \ref{algorithm1}, in each iterative step, the computational complexity of updating $U$ is $\mathcal{O}(nc\tilde{d})$, which is the same as that of updating $V$ and $\gamma$. Then the computational complexity of updating $\tilde{X}$ focuses on performing eigenvalue decomposition, which is $\mathcal{O}(n^3)$. Therefore, the overall computational complexity of the proposed AFCM method in Algorithm \ref{algorithm1} is $\mathcal{O}(tn^3 + 3tnc\tilde{d})$, which can be simplified to $\mathcal{O}(tn^3)$.
For Algorithm \ref{algorithm2}, since $\tilde{X}$ is not updated, the algorithm operates in the original dimension $d$, resulting in a total computational complexity of $\mathcal{O}(tncd)$.

\begin{table}[h]
\centering
\caption{Description of data sets.}
\label{data}
\begin{tabular}{ccccc}
\toprule
Data set       & Samples & Dimensions & Clusters           \\
\midrule
Iris           & 150     & 4          & 3  \\
Breast         & 699     & 10         & 2  \\
Vehicle        & 846     & 18         & 4 \\
UPPS           & 1854    & 256        & 10 \\
Jaffe50        & 213     & 1024       & 10  \\
warpPIE10P     & 210     & 2420       & 10 \\
Olivetti       & 900     & 2500       & 10 \\
ORL64x64       & 400     & 4096       & 40  \\
Pose07         & 1629    & 4096       & 68 \\
Pose29         & 1632    & 4096       & 68 \\
\bottomrule
\end{tabular}
\end{table}

\section{Experiment}

\subsection{Description of Datasets}
In this subsection, we provide a description of the experimental datasets. Ten real-world datasets are used in experiments, including Iris\footnote{https://archive.ics.uci.edu}, Breast\footnotemark[1], Vehicle\footnotemark[1], USPS\footnote{http://www.cad.zju.edu.cn/home/dengcai/Data/MLData.html}, Jaffe50\footnote{https://www.kaggle.com/}, warpPIE10P\footnote{https://jundongl.github.io/scikit-feature/datasets.html}, Olivetti\footnotemark[1], ORL64x64\footnotemark[2], Pose07\footnote{https://www.ri.cmu.edu/publications/the-cmu-pose-illumination-and-expression-pie-database/}, and Pose29\footnotemark[5]. The details of these real-world datasets are given in Table \ref{data}.

\begin{table*}[t]
\caption{Clustering results for AFCM on ten real-world data sets (average ACC, NMI and ARI $\pm$ standard deviations$\times$1E+2). The bold characters are the optimal results.}
{\setlength{\tabcolsep}{3.5pt} 
\begin{tabular*}{\hsize}{@{\extracolsep\fill}cccccccccc}
\toprule
Dataset                                     &Metric           &FCM                          &FCM-ER                       &SC                           &RSFKM                        &CDKM                         &FKPS                         &FCAG                         & AFCM\\
\midrule
\multirow{3}{*}{Iris}                       &ACC              &88.73($\pm$0.20)             &89.33($\pm$0.00)             &94.67($\pm$0.00)             &88.67($\pm$0.00)             &88.67($\pm$2.58)             &92.00($\pm$0.00)             &95.47($\pm$2.40)             &\bf{96.13($\pm$1.60)}\\
                                            &NMI              &73.01($\pm$0.37)             &73.64($\pm$0.00)             &85.88($\pm$0.00)             &75.48($\pm$0.63)             &73.64($\pm$0.00)             &79.82($\pm$0.00)             &86.32($\pm$3.36)             &\bf{87.49($\pm$2.89)}\\
                                            &ARI              &71.64($\pm$0.41)             &72.78($\pm$0.00)             &78.29($\pm$0.02)             &75.60($\pm$0.00)             &71.63($\pm$0.00)             &78.74($\pm$0.00)             &87.48($\pm$5.78)             &\bf{89.07($\pm$3.93)}\\
\midrule
\multirow{3}{*}{Breast}                     &ACC              &94.99($\pm$0.00)             &96.28($\pm$0.00)             &96.04($\pm$0.00)             &95.28($\pm$0.00)             &95.28($\pm$0.00)             &96.28($\pm$0.00)             &95.99($\pm$0.00)             &\bf{96.57($\pm$0.00)}\\
                                            &NMI              &69.15($\pm$0.00)             &75.60($\pm$0.00)             &74.36($\pm$4.55)             &70.53($\pm$0.00)             &70.49($\pm$0.00)             &75.76($\pm$0.00)             &74.17($\pm$0.00)             &\bf{78.00($\pm$0.00)}\\
                                            &ARI              &80.74($\pm$0.28)             &85.54($\pm$0.15)             &84.63($\pm$0.00)             &81.79($\pm$0.00)             &81.79($\pm$0.00)             &85.60($\pm$0.00)             &84.46($\pm$0.00)             &\bf{86.64($\pm$0.00)}\\
\midrule
\multirow{3}{*}{Vehicle}                    &ACC              &37.12($\pm$0.13)             &36.90($\pm$0.51)             &41.80($\pm$0.57)             &38.09($\pm$1.69)             &36.87($\pm$0.52)             &38.75($\pm$0.76)             &42.08($\pm$3.56)             &\bf46.74($\pm$2.83)\\
                                            &NMI              &10.53($\pm$1.58)             &10.60($\pm$1.70)             &16.15($\pm$2.42)             &10.59($\pm$0.75)             &10.59($\pm$1.71)             &11.79($\pm$2.76)             &14.63($\pm$4.44)             &\bf19.81($\pm$1.82)\\
                                            &ARI              &7.86($\pm$0.83)              &8.36($\pm$0.21)              &11.15($\pm$0.01)             &9.11($\pm$0.37)              &7.88($\pm$0.87)              &8.55($\pm$0.00)              &11.17($\pm$2.27)             &\bf15.75($\pm$2.25)\\
\midrule
\multirow{3}{*}{USPS}                       &ACC              &65.78($\pm$1.56)             &64.53($\pm$1.42)             &69.32($\pm$2.48)             &66.04($\pm$3.59)             &66.59($\pm$2.67)             &66.42($\pm$2.28)             &\bf74.02($\pm$2.48)          &72.07($\pm$4.32)\\
                                            &NMI              &62.91($\pm$0.91)             &62.72($\pm$0.04)             &73.61($\pm$1.52)             &64.00($\pm$1.33)             &62.67($\pm$1.28)             &63.80($\pm$1.60)             &71.17($\pm$1.08)             &\bf77.59($\pm$1.58)\\
                                            &ARI              &54.11($\pm$1.38)             &53.53($\pm$1.32)             &62.83($\pm$2.99)             &55.04($\pm$3.13)             &53.94($\pm$1.79)             &55.27($\pm$2.00)             &62.85($\pm$1.15)             &\bf65.67($\pm$4.21)\\
\midrule
\multirow{3}{*}{Jaffe50}                    &ACC              &92.77($\pm$0.00)             &84.55($\pm$3.30)             &86.43($\pm$2.96)             &86.90($\pm$3.60)             &77.65($\pm$6.64)             &42.18($\pm$2.46)             &89.95($\pm$3.57)             &\bf94.93($\pm$3.86)\\
                                            &NMI              &92.27($\pm$0.00)             &89.75($\pm$2.70)             &90.38($\pm$1.34)             &89.47($\pm$2.76)             &85.76($\pm$3.28)             &48.00($\pm$2.77)             &92.33($\pm$2.14)             &\bf95.20($\pm$2.15)\\
                                            &ARI              &85.88($\pm$0.00)             &79.78($\pm$5.15)             &82.09($\pm$0.00)             &79.67($\pm$2.50)             &72.09($\pm$6.00)             &49.09($\pm$2.87)             &84.96($\pm$4.589)            &\bf91.23($\pm$4.26)\\
\midrule
\multirow{3}{*}{warpPIE10P}                 &ACC              &28.24($\pm$1.76)             &62.00($\pm$5.52)             &48.48($\pm$1.62)             &27.95($\pm$1.15)             &28.52($\pm$1.29)             &29.33($\pm$1.78)             &44.00($\pm$3.42)             &\bf50.57($\pm$0.29)\\
                                            &NMI              &30.06($\pm$2.13)             &29.91($\pm$3.49)             &61.05($\pm$2.81)             &30.07($\pm$1.67)             &31.01($\pm$1.63)             &31.65($\pm$2.75)             &52.06($\pm$3.05)             &\bf63.79($\pm$0.86)\\
                                            &ARI              &9.53($\pm$1.98)              &9.16($\pm$1.27)              &35.80($\pm$3.77)             &9.27($\pm$1.27)              &9.85($\pm$1.46)              &10.67($\pm$1.45)             &26.81($\pm$3.45)             &\bf39.11($\pm$1.02)\\
\midrule
\multirow{3}{*}{Olivetti}                   &ACC              &53.88($\pm$3.24)             &49.76($\pm$3.18)             &68.56($\pm$0.00)             &51.03($\pm$3.47)             &47.10($\pm$3.20)             &51.90($\pm$5.01)             &59.11($\pm$2.55)             &\bf70.41($\pm$2.89)\\
                                            &NMI              &52.10($\pm$2.38)             &50.98($\pm$3.66)             &71.88($\pm$0.37)             &51.89($\pm$1.94)             &50.93($\pm$2.84)             &53.75($\pm$3.78)             &60.71($\pm$4.49)             &\bf74.98($\pm$1.68)\\
                                            &ARI              &36.92($\pm$2.61)             &33.45($\pm$2.89)             &50.66($\pm$1.06)             &34.82($\pm$2.09)             &33.32($\pm$2.72)             &36.19($\pm$3.60)             &43.56($\pm$5.48)             &\bf59.92($\pm$2.23)\\
\midrule
\multirow{3}{*}{ORL64x64}                   &ACC              &59.40($\pm$1.39)             &57.98($\pm$1.65)             &58.55($\pm$2.36)             &58.68($\pm$2.44)             &58.10($\pm$2.79)             &58.58($\pm$2.22)             &62.38($\pm$1.35)             &\bf62.90($\pm$2.65)\\
                                            &NMI              &77.63($\pm$0.99)             &77.46($\pm$0.79)             &77.69($\pm$1.94)             &77.67($\pm$0.88)             &76.82($\pm$1.38)             &77.74($\pm$1.02)             &\bf81.75($\pm$0.88)          &80.68($\pm$0.46)\\
                                            &ARI              &46.00($\pm$1.25)             &45.29($\pm$2.05)             &41.64($\pm$5.30)             &45.956($\pm$2.20)            &44.09($\pm$2.82)             &46.10($\pm$1.97)             &\bf53.94($\pm$2.14)          &50.94($\pm$1.58)\\

\midrule
\multirow{3}{*}{Pose07}                     &ACC              &15.95($\pm$0.44)             &15.49($\pm$0.27)             &28.86($\pm$0.83)             &15.02($\pm$0.34)             &16.24($\pm$0.67)             &15.43($\pm$0.48)             &23.20($\pm$0.77)             &\bf32.42($\pm$1.07)\\
                                            &NMI              &42.86($\pm$0.36)             &42.28($\pm$0.22)             &53.54($\pm$1.24)             &42.04($\pm$0.45)             &42.23($\pm$0.64)             &42.10($\pm$0.48)             &54.34($\pm$0.43)             &\bf60.54($\pm$0.62)\\
                                            &ARI              &8.80($\pm$0.31)              &5.84($\pm$0.22)              &11.26($\pm$1.24)             &5.25($\pm$0.21)              &5.84($\pm$0.44)              &5.49($\pm$0.37)              &12.52($\pm$0.54)             &\bf18.58($\pm$1.31)\\
\midrule
\multirow{3}{*}{Pose29}                     &ACC              &17.12($\pm$0.29)             &15.40($\pm$0.37)             &26.65($\pm$0.42)             &15.92($\pm$0.55)             &17.46($\pm$0.53)             &15.70($\pm$0.43)             &25.03($\pm$0.65)             &\bf32.61($\pm$1.85)\\
                                            &NMI              &44.01($\pm$0.47)             &42.72($\pm$0.63)             &43.49($\pm$2.01)             &43.59($\pm$0.31)             &42.25($\pm$0.64)             &41.35($\pm$0.47)             &55.69($\pm$0.30)             &\bf59.15($\pm$1.11)\\
                                            &ARI              &6.89($\pm$0.25)              &6.05($\pm$0.32)              &4.69($\pm$1.10)              &6.21($\pm$0.33)              &6.36($\pm$0.44)              &5.72($\pm$0.38)              &13.98($\pm$0.38)             &\bf17.51($\pm$1.46)\\
\bottomrule
\end{tabular*}}
\label{table:result1}
\end{table*}

\subsection{Experimental Settings}

In this experiment, seven state-of-the-art clustering algorithms are selected as the comparative algorithms to compared with the proposed AFCM algorithm. These comparative algorithms include Fuzzy C-Means (FCM) \cite{bezdek2013pattern}, Fuzzy C-Means with Entropy Regularization (FCM-ER) \cite{li1995maximum}, Spectral Clustering (SC) \cite{von2007tutorial}, Robust Sparse Fuzzy K-Means (RSFKM) \cite{xu2016robust}, Coordinate Descent K-Means (CDKM) \cite{9444882}, Fuzzy K-Means with Pattern Shrinking (FKPS) \cite{zhao2021robust} and Fast Clustering model with Anchor Guidance (FCAG)\cite{nie2023fast}.

To ensure a consistent standardization of the data, each dataset is subjected to normalization using the min-max normalization method. For fair comparison with the competitors, based on their given parameter lists, we use the grid-search method to select the best parameter values for each comparative algorithm. There are two parameters in the proposed AFCM algorithm including the number of the nearest neighbors $k$ in the normalized Laplacian matrix and the regularization parameter $\lambda$, and their parameter lists are given as $[3, 4, 5, 6, 8, 10, 12]$ and [1e-1, 1e1, 1e2, 1e3, 1e4, 1e5, 1e6], respectively. Similar to the competitors, we also use the grid-search method to select parameters for AFCM.
Accuracy (ACC), Normalized Mutual Information (NMI), and Adjusted Rand Index (ARI) are three commonly utilized metrics for assessing the performance of clustering algorithms \cite{6832486}, and they are employed in our paper as well.

\subsection{Visualization Experiments on Toy Datasets}
We conduct visualization experiments on two synthetic toy datasets to demonstrate the capability of the proposed model on handling data with non-Gaussian clusters. The first one consists of two spiral-shaped clusters, with each cluster containing 500 samples. The second one consists of three ring-shaped clusters, with each cluster containing 300 samples.

First, we perform FCM on the toy datasets and the clustering results are shown in Fig. \ref{fig2a} and Fig. \ref{fig2b}. Obviously, FCM cannot handle this type of data. Then, we show the clustering results of the degenerate AFCM in Fig. \ref{fig2c} and Fig. \ref{fig2d}. Since the degenerate AFCM performs clustering directly in the original manifold, it consequently fails as well. Finally, we present the clustering results of the proposed AFCM that performs clustering in the learned manifold. In Fig. \ref{fig2e} and Fig. \ref{fig2g}, we show the clustering results of AFCM in the original manifold. In Fig. \ref{fig2f} and Fig. \ref{fig2h}, we show the clustering results of AFCM in the learned manifold.
By observing the clustering results, it is clear that the proposed AFCM successfully projected the data from a non-Gaussian manifold into a Gaussian manifold, and then successfully clustered the data.

\subsection{Evaluation on Real-World Datasets}
In order to further verify the effectiveness of the proposed AFCM method, we compare it with seven state-of-the-art clustering methods on ten real-world datasets. To ensure a fair comparison, we execute each algorithm ten times and present the average results along with standard deviations. The clustering outcomes are evaluated using three frequently used metrics: ACC, NMI, and ARI. The clustering results are shown in Table \ref{table:result1}.

According to the clustering results in Table \ref{table:result1}, the proposed AFCM obtains the optimal results on most datasets, except for the USPS and ORL64x64 datasets. The comparative algorithm FCAG exhibits better performance than the proposed AFCM on the USPS in terms of ACC, but AFCM outperforms FCAG in terms of NMI and ARI on the same dataset. A similar situation also occurs with the ORL64x64 dataset. Therefore, while the performance of AFCM is comparable to FCAG on the USPS and ORL64x64 datasets, it obtains the best clustering results on the other eight datasets, validating the effectiveness of the proposed AFCM method. Moreover, it is observed that the graph based method AFCM, SC, and FCAG usually perform better than other non-graph based methods. This observation indicates that graph embedding is an excellent method to help improve clustering performance.

\subsection{Ablation Experiments}
In this subsection, we want to verify two facts through the ablation experiment. The first one is that the parameter-free degenerate AFCM algorithm can be an excellent alternative to K-Means. The second one is that simultaneously performing clustering and manifold learning may obtain better results than performing them separately.

\begin{table*}[t]
\caption{Clustering results for the ablation experiments on eight real-world data sets (average ACC, NMI and ARI $\pm$ standard deviations$\times$1E+2). The bold characters are the optimal results. Ablation-1 and Ablation-2 adopt the two-stage strategy, whereas AFCM takes the one-stage strategy.}
{\setlength{\tabcolsep}{2.8pt} 
\begin{tabular*}{\hsize}{@{\extracolsep\fill}cccccccccc}
\toprule
Metric                       &Method           &Iris                         &Breast                       &Vehicle                      &USPS                         &warpPIE10P                   &Olivetti                     &ORL64x64                     &Pose07   \\
\midrule
\multirow{3}{*}{ACC}         &Ablation-1       &94.67($\pm$0.00)             &95.85($\pm$0.00)             &41.76($\pm$0.46)             &69.31($\pm$3.34)             &47.52($\pm$2.65)             &67.69($\pm$0.16)             &59.45($\pm$1.83)             &27.26($\pm$1.38)\\
                             &Ablation-2       &94.67($\pm$0.00)             &95.85($\pm$0.00)             &44.65($\pm$2.51)             &71.65($\pm$3.89)             &50.29($\pm$0.32)             &68.54($\pm$2.10)             &61.63($\pm$1.93)             &31.33($\pm$1.05)\\
                             &AFCM             &\bf96.13($\pm$1.60)          &\bf96.57($\pm$0.00)          &\bf46.74($\pm$2.83)          &\bf72.07($\pm$4.32)          &\bf50.57($\pm$0.29)          &\bf70.41($\pm$2.89)          &\bf62.90($\pm$2.65)          &\bf{32.42($\pm$1.07)}\\
\midrule
\multirow{3}{*}{NMI}         &Ablation-1       &85.88($\pm$0.00)             &73.45($\pm$0.00)             &16.08($\pm$0.00)             &75.04($\pm$2.46)             &58.90($\pm$3.17)             &70.69($\pm$1.59)             &76.98($\pm$1.62)             &52.64($\pm$1.84)\\
                             &Ablation-2       &85.88($\pm$0.00)             &73.45($\pm$0.00)             &18.01($\pm$1.21)             &76.39($\pm$2.43)             &62.52($\pm$1.27)             &71.72($\pm$2.82)             &\bf80.72($\pm$0.78)          &59.93($\pm$0.76)\\
                             &AFCM             &\bf87.49($\pm$2.89)          &\bf78.00($\pm$0.00)          &\bf19.81($\pm$1.82)          &\bf77.59($\pm$1.58)          &\bf63.79($\pm$0.86)          &\bf74.98($\pm$1.68)          &80.68($\pm$0.46)             &\bf{60.54($\pm$0.62)}\\
\midrule
\multirow{3}{*}{ARI}         &Ablation-1       &85.15($\pm$0.00)             &83.94($\pm$0.00)             &10.86($\pm$0.00)             &61.85($\pm$5.43)             &32.04($\pm$3.28)             &48.86($\pm$1.45)             &40.89($\pm$3.19)             &10.84($\pm$1.40)\\
                             &Ablation-2       &85.15($\pm$0.00)             &83.94($\pm$0.00)             &14.34($\pm$2.01)             &64.00($\pm$5.42)             &37.59($\pm$1.40)             &56.75($\pm$3.82)             &50.41($\pm$1.65)             &16.76($\pm$1.59)\\
                             &AFCM             &\bf89.07($\pm$3.93)          &\bf86.64($\pm$0.00)          &\bf15.75($\pm$2.25)          &\bf66.57($\pm$4.21)          &\bf39.11($\pm$1.02)          &\bf59.92($\pm$2.23)          &\bf50.94($\pm$1.58)          &\bf18.58($\pm$1.31)\\
\bottomrule
\end{tabular*}}
\label{table:result2}
\end{table*}

\begin{figure*}[t]
\centering
\subfloat[Iris]{\includegraphics[width=1.8in]{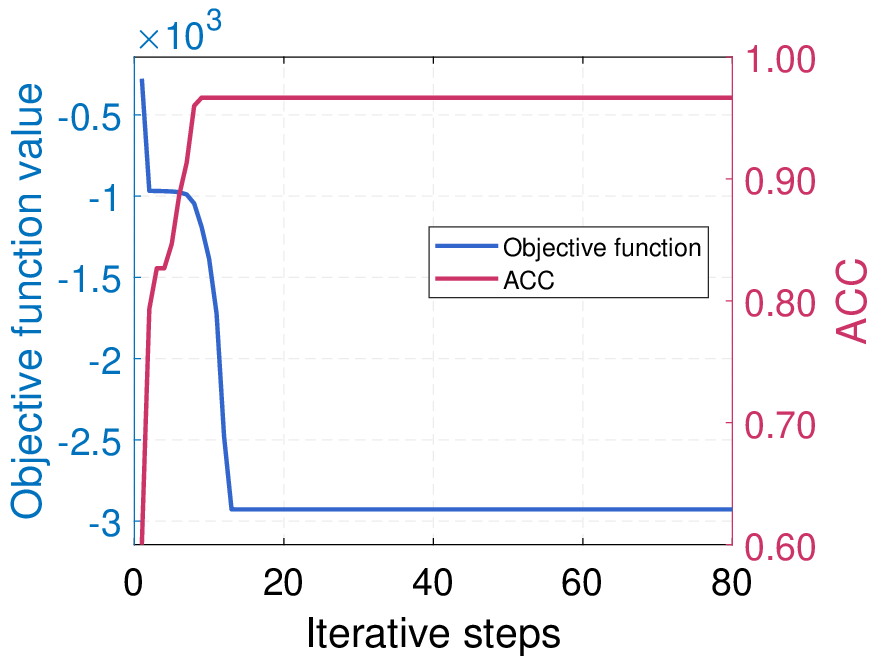}%
\label{fig3a}}
\subfloat[Breast]{\includegraphics[width=1.8in]{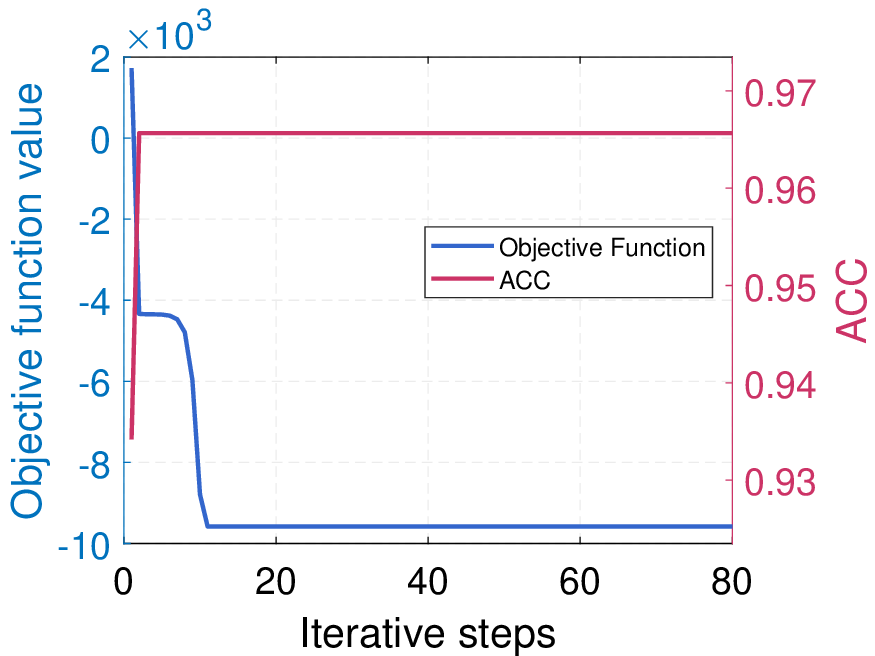}%
\label{fig3b}}
\subfloat[Vehicle]{\includegraphics[width=1.8in]{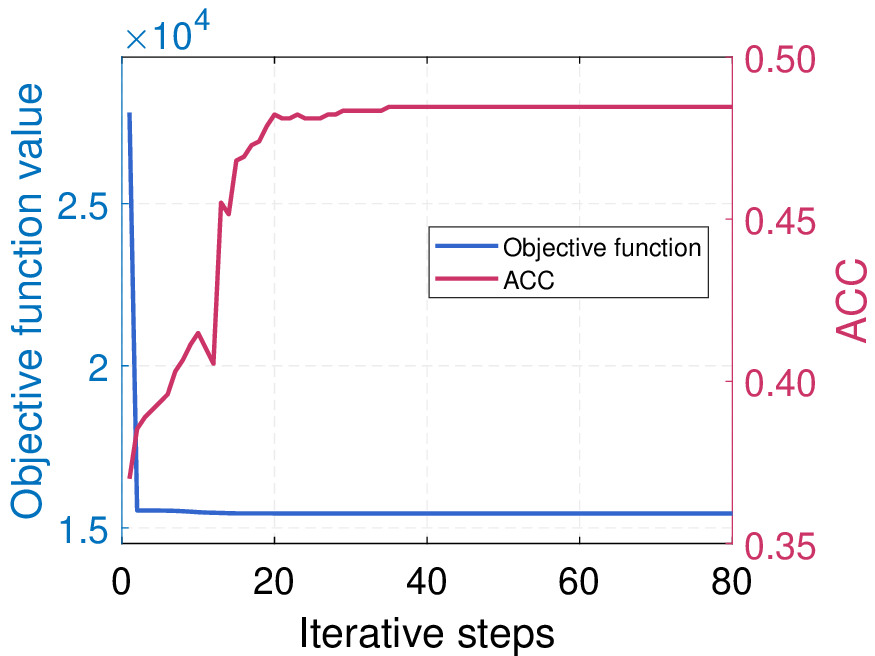}%
\label{fig3c}}
\subfloat[USPS]{\includegraphics[width=1.8in]{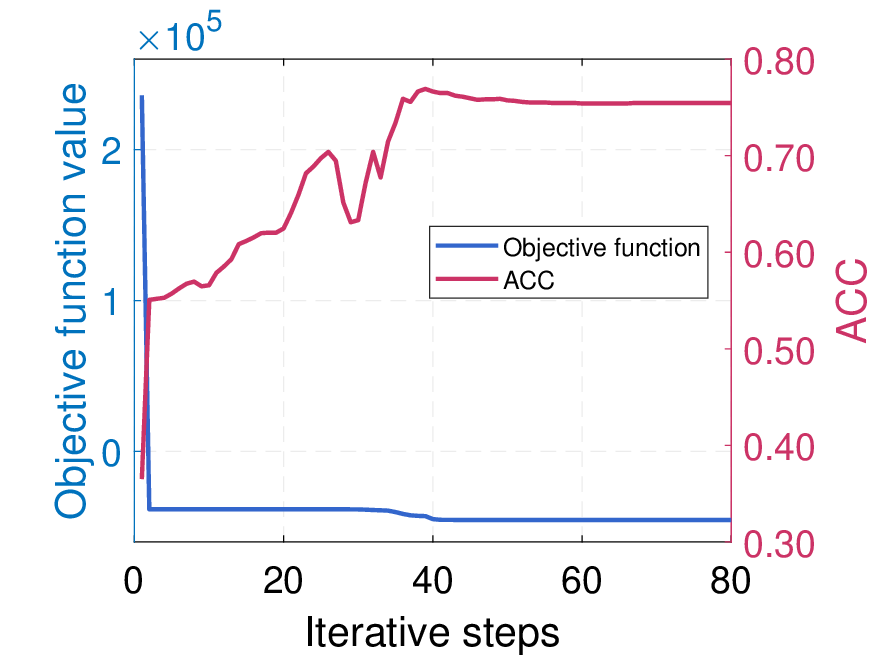}%
\label{fig3d}}
\\
\subfloat[warpPIE10P]{\includegraphics[width=1.8in]{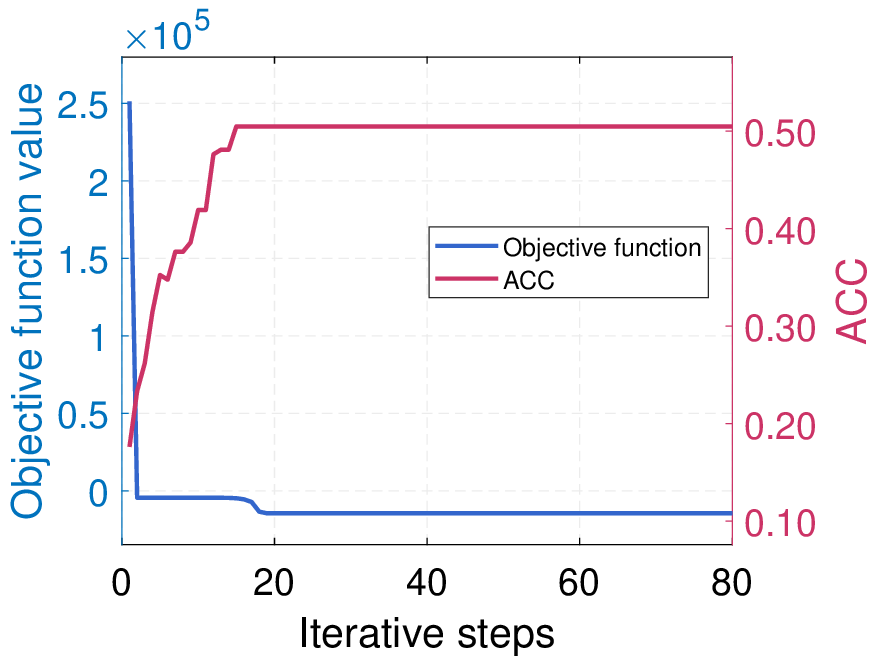}%
\label{fig3e}}
\subfloat[Olivetti]{\includegraphics[width=1.8in]{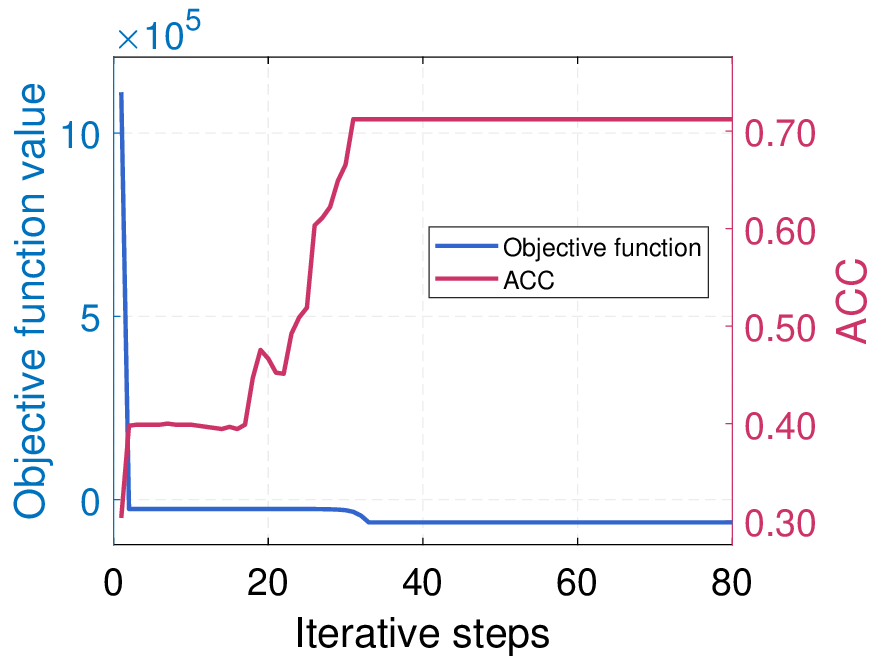}%
\label{fig3f}}
\subfloat[ORL64x64]{\includegraphics[width=1.8in]{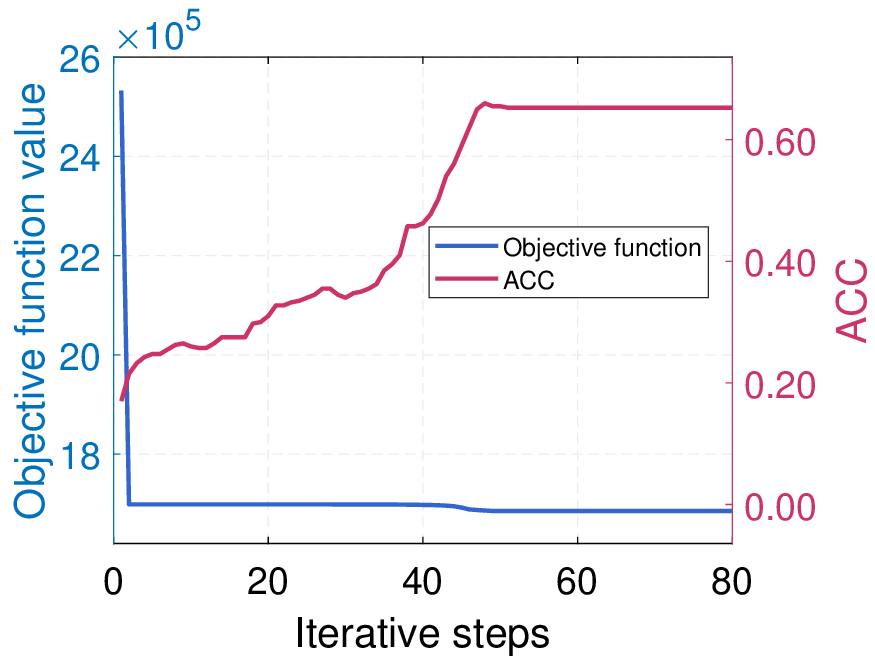}%
\label{fig3g}}
\subfloat[Pose07]{\includegraphics[width=1.8in]{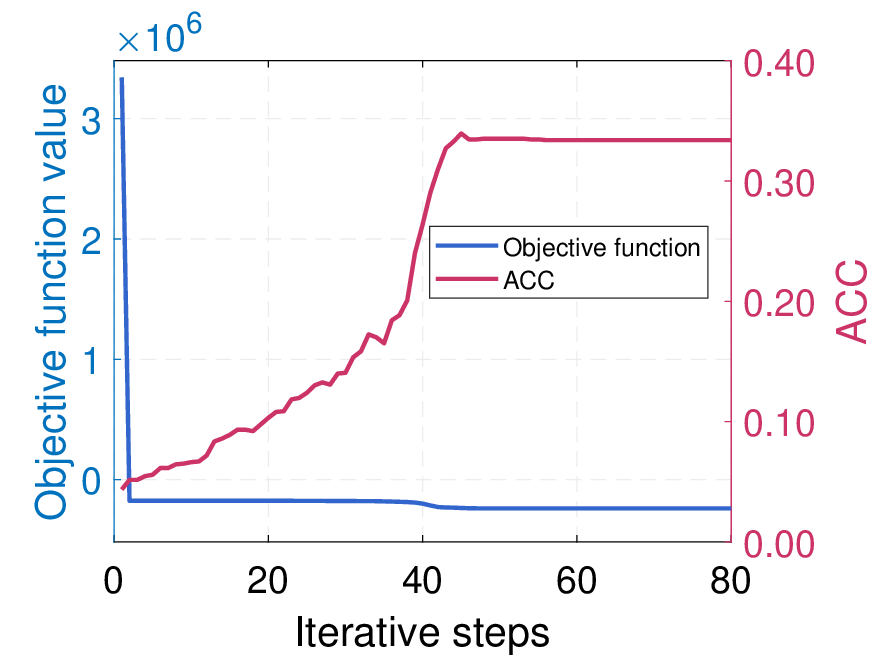}%
\label{fig3h}}
\caption{Convergence and ACC curves for AFCM on eight real-world datasets.}
\label{fig3}
\end{figure*}

To verify this, we construct two comparative algorithms named as Ablation-1 and Ablation-2.
Ablation-1 is a two-stage method, which first computes the $c$ minimum eigenvectors of the normalized Laplacian matrix $\hat{L}$ and then performs K-Means on these eigenvectors. Ablation-2 is also a two-stage method, which first computes the $c$ minimum eigenvectors of the normalized Laplacian matrix $\hat{L}$  and then performs the degenerate AFCM on the eigenvectors. We run each algorithm ten times, recording the average clustering results, which are presented in Table \ref{table:result2}.


According to the results in Table \ref{table:result2}, it is observed that the performance of Ablation-2 is better than that of Ablation-1, which means that the performance of the degenerate AFCM is  better than K-means. The degenerate AFCM has the similar computational complexity with K-Means but exhibit a better performance than K-Means. Therefore it is an excellent alternative to K-means. Furthermore, the performance of the one-stage method AFCM is better than Ablation-1 and Ablation-2 that adopts two-stage approaches, which means perform clustering and manifold learning together is better than performing them separately.

\vspace{-0.15cm}
\subsection{Convergence Experiments}
In order to validate the convergence of the proposed AFCM algorithm, we present the curves of the objective function values and the ACC values in Figure \ref{fig3}. The experiment was conducted using eight real-world datasets, with the maximum number of iterations set to 80.

From the results depicted in Figure \ref{fig3}, it is evident that the curves of the objective function values for all eight datasets exhibit a monotonically decreasing trend. This observation confirms the efficacy of the optimization strategy employed in the AFCM model.
Additionally, the ACC curves in Figure \ref{fig3} show a general trend of increasing with the iterative steps, albeit not strictly monotonic. There are instances where the ACC values may decrease temporarily; however, upon convergence, the ACC values significantly surpass those at the initial iterative step. This pattern validates the rationality and effectiveness of the proposed AFCM model design.

Overall, the experimental results presented in Figure \ref{fig3} demonstrate the successful convergence and effectiveness of the AFCM algorithm across diverse real-world datasets.

\section{Conclusions}
In this paper, we propose a novel Adaptive Fuzzy C-Means with graph embedding model (AFCM) that can automatically learn the membership degree hyper-parameter and handle data with non-Gaussian clusters. By establishing the equivalent connections between FCM with entropy regularization and the generalized Gaussian mixture model, we reveal that the entropy regularization hyper-parameter in FCM  can be interpreted as the scale parameter in the generalized Gaussian mixture model. Therefore, the entropy regularization hyper-parameter can be learned in a similar way to the scale parameter in the mixture model. Furthermore, by introducing the graph embedding regularization, the proposed AFCM model is capable of handling data with non-Gaussian clusters. Note that, the proposed method perform clustering and manifold learning simultaneously, which is better than performing them separately, and we validate this by experiments. Moreover, the proposed AFCM can degenerate into a parameter-free FCM. This simplified AFCM variant maintains a computational complexity of $\mathcal{O}(ncd)$, comparable to K-Means, yet outperforms K-Means in performance. However, while the proposed AFCM achieves good performance, it still relies on the original FCM to conduct clustering in the newly learned manifold. In the future, we aim to integrate more advanced techniques into AFCM to enhance its clustering performance in the new manifold.


{\appendix
\section*{Proof for proposition \ref{pro1}} \label{pro1_proof}
In this section, we presents a rigorous mathematical proof of Proposition \ref{pro1}. To facilitate the readers, we rewritten the equivalent objective function of the generalized Gaussian mixture model as follows.
\begin{equation}\label{pro1_ea1}
\hspace{-0.2cm}
\begin{gathered}
\mathop {\min }\limits_{U,\alpha, V, \Sigma, \beta ,m} \mathop \sum \limits_{i = 1}^n \mathop \sum \limits_{j = 1}^c {u_{ij}}\bigg\{ m{{ {{\left[ {{{(x_i  - {v_j})}^T}{\Sigma_j^{ - 1}}(x_i - {v_j})} \right]}} }^\beta } \\
\quad + \frac{1}{2}\log |{\Sigma _j}| - \log {\alpha _j} - \log \frac{{\beta \Gamma (\frac{d}{2}){m^{\frac{d}{{2\beta }}}}}}{{{\pi ^{\frac{d}{2}}}\Gamma (\frac{d}{{2\beta }})}} + \log {u_{ij}} \bigg\}, \\
\ \ {\rm s.t.} \ \ \mathop \sum \limits_{j = 1}^c {\alpha _j} = 1, \ \ 0 < {\alpha _j} < 1,  \\
\ \ \ \ \ \ \ \ \ \mathop \sum \limits_{j = 1}^c {u_{ij}} = 1, \ \ 0 < {u_{ij}} < 1
\end{gathered}
\end{equation}
where $u_{ij}$ is the membership degree denoting the probability of sample $x_i$ being assigned to the $j$-th cluster, $\alpha  = \{ {\alpha _1},{\alpha _2},..,{\alpha _c}\}$, $V = \{ {v_1},{v_2},..,{v_c}\} $ and $\Sigma  = \{ {\Sigma _1},{\Sigma _2},..,{\Sigma _c}\} $, denote the sets of mixing coefficients, means and scale matrices, and $\beta$ and $m$ denotes the shape parameter and the scale parameter in the generalized Gaussian mixture model.

We will demonstrate that the update equations for the equivalent objective function outlined in Eq. (\ref{pro1_ea1}) coincide with those derived for the log-likelihood function of the generalized Gaussian mixture model.

\subsection{The maximum log-likelihood function}

Let $x\in \mathbb{R}^d$ be a $d$-dimensional random vector, and the probability density function of the generalized Gaussian distribution is given as below.
\begin{equation}\label{ea2}
\begin{gathered}
g(x|\theta )\! =\!  \frac{{\beta \Gamma (\frac{d}{2}){m^{\frac{d}{{2\beta }}}}}}{{{\pi ^{\frac{d}{2}}}\Gamma (\frac{d}{{2\beta }})}}  |\Sigma {|^{ - \frac{1}{2}}} \exp\! \left\{\!  - m  {{\left[ {{{(x\! - \!v)}^T}{\Sigma ^{ - 1}}(x\! - \!v)} \right]}^\beta } \! \right\} \\
\end{gathered}
\end{equation}
where $\theta = \{v,\Sigma,\beta,m\}$ denotes the set of parameters, $v\in\mathbb{R}^d$ denotes the mean, $\Sigma\in\mathbb{R}^{d\times d}$ denotes the positive definite covariance matrix, $\beta\in\mathbb{R}^+$ denotes the shape parameter, and $m$ denotes the scale parameter.

Based on the probability density function the generalized Gaussian distribution depicted in Eq. (\ref{ea2}), the probability density function of the generalized Gaussian mixture model can be defined as
\begin{equation}\label{ea3}
\begin{gathered}
f(x|\Theta ) = \mathop \sum \limits_{j = 1}^c {\alpha _j}g(x|{\theta _j}), \ \ {\rm s.t.}\sum_{j=1}^{c}\alpha_j=1
\end{gathered}
\end{equation}
where $\Theta  = \{ \alpha ,V,\Sigma ,\beta ,m\} $ denotes the set of all parameters in the generalized Gaussain mixture model, ${\theta _j} = \left\{ {{v_j},{\Sigma _j},\beta ,m} \right\}$ denotes the parameter set of the $j$-th component, $\alpha  = \{ {\alpha _1},{\alpha _2},..,{\alpha _c}\}$ denotes the set of mixing coefficients, and $c$ denotes the number of components.

Then the expression of the log-likelihood function for the generalized Gaussian mixture model is presented as follows.
\begin{equation}\label{ea4}
\begin{gathered}
  L(\Theta |X) \! \! =\! \! \mathop \sum \limits_{i = 1}^n \!  \log f({x_i}|\Theta ) \! = \!\! \mathop \sum \limits_{i = 1}^n \!\log\! \mathop \sum \limits_{j = 1}^c \! {\alpha _j}g({x_i}|{\theta _j}), \\
   {\rm s.t. \ \ }\sum_{j=1}^{c}\alpha_j=1
\end{gathered}
\end{equation}

\subsection{EM for problem in (\ref{ea4})}
Owing to the combination of logarithm and sum within the problem depicted in Eq. (\ref{ea4}), direct optimization becomes challenging, necessitating the utilization of the EM algorithm. There are two main steps in the EM, including E-step and M-step.

{\bf In E-step:} According to the log-likelihood function in Eq. (\ref{ea4}) one has
\begin{equation}\label{ea5}
\begin{aligned}
p(j|{x_i},{\Theta ^{t - 1}}) \!=\! \frac{{\alpha _j^{t - 1}g({x_i}|\theta _j^{t - 1})}}{{\mathop \sum \limits_{j = 1}^c \alpha _j^{t - 1}g({x_i}|\theta _j^{t - 1})}} \qquad\qquad\qquad
\\\! = \! \frac{{{\alpha _j^{t - 1}}|{\Sigma _j^{t - 1}} \! {|^{ - \frac{1}{2}}}\! \exp \left\{ \! { - m^{t - 1}(d_{ij})^{t-1}} \! \right\}}}{{\mathop \sum \limits_{j = 1}^c \!\! {{\alpha _j^{t - 1}}|{\Sigma _j^{t - 1}}\!{|^{ - \frac{1}{2}}} \! \exp \left\{ \! { - m^{t - 1} (d_{ij})^{t-1} } \! \right\}}}}
\end{aligned}
\end{equation}
where $(d_{ij})^{t-1}= {{ { {{{\left[ {{{(x_i \! - \!{v_j^{t-1}})}^T}{(\Sigma_j^{t-1}) ^{ - 1}}(x_i\! -\! {v_j^{t-1}})} \right]}}} } }^{\beta^{t - 1}} }$; symbols $\Theta ^{t - 1}$ and $\theta^{t-1}_j$ denote the parameter sets estimated in last iteration, and $p(j|{x_i},{\Theta ^{t - 1}})$ can be considered as the posterior probability of the $i$-th sample being generated by the $j$-th mixture component.

{\bf In M-step:} Fix $p(j|{x_i},{\Theta ^{t - 1}})$ and optimize the following problem.
\begin{equation}\label{ea6}
\begin{aligned}
&\mathop {\max }\limits_{\Theta } Q(\Theta , \! {\Theta ^{t \! - \! 1}} \! ) \!=\! \mathop {\max }\limits_{\Theta } \! \sum\limits_{i = 1}^n \sum\limits_{j = 1}^c {p(j\left| {{x_i},{\Theta ^{t \! - \! 1}})} \! \right.\log {\alpha _j}g({x_i}\left| {{\theta _j})} \right.}\\
= &\mathop {\max }\limits_{\Theta } \sum\limits_{i = 1}^n \sum\limits_{j = 1}^c p(j\left| {{x_i},{\Theta ^{t - 1}})} \right. \log \bigg \{  \alpha_j  { \frac{{\beta \Gamma (\frac{d}{2}){m^{\frac{d}{{2\beta }}}}}}{{{\pi ^{\frac{d}{2}}}\Gamma (\frac{d}{{2\beta }})}}}  \times \\
&\qquad \ \ |\Sigma_j|^{\frac{-1}{2}} \exp\left [ -m   \left( (x_i-v_j)^T\Sigma_j^{-1}(x_i-v_j)\right)^\beta \right] \bigg \} \\
\Leftrightarrow &\mathop {\min }\limits_\Theta  \sum\limits_{i = 1}^n  \sum\limits_{j = 1}^c  p(j  \left| {{x_i},{\Theta ^{t - 1}})} \right.  \bigg\{  m{{\left[ {{{{(x_i  - {v_j})}^T}{\Sigma_j^{ - 1}} (x_i - {v_j})}} \right]} ^\beta } \\
&\qquad \ \  + \frac{1}{2}\log |{\Sigma _j}| - \log {\alpha _j} - \log { \frac{{\beta \Gamma (\frac{d}{2}){m^{\frac{d}{{2\beta }}}}}}{{{\pi ^{\frac{d}{2}}}\Gamma (\frac{d}{{2\beta }})}} } \bigg\} \\
& \qquad\qquad\quad \ {\rm s.t.}\ \ \mathop \sum \limits_{j = 1}^c {\alpha _j} = 1, \ \ 0 < {\alpha _j} < 1,\ \
\end{aligned}
\end{equation}
where $\Theta  = \{ \alpha ,V,\Sigma ,\beta ,m\}$.

Through iterative execution of the E-step and M-step until converging, one can obtain a suitable local minimum for the log-likelihood function presented in Eq. (\ref{ea4}).

\subsection{Proof for equivalence}
The proof for the equivalence between the objective function in Eq. (\ref{pro1_ea1}) and the log-likelihood function in Eq. (\ref{ea4}) is mainly divided into two parts. In the first part, we prove that $u_{ij}$ in Eq. (\ref{pro1_ea1}) have the same update equation with $p(j|{x_i},{\Theta ^{t - 1}})$ in Eq. (\ref{ea5}), where $u_{ij}$ denotes the membership degree and $p(j|{x_i},{\Theta ^{t - 1}})$ is the posterior probability derived from the E-step. In the second part, we prove that the equivalent objective function in Eq. (\ref{pro1_ea1}) with $U$ being fixed has the same form with the Q function in Eq. (\ref{ea6}).

{\bf In the first part:} When we update $U$ with the other variables being fixed, the objective function in Eq. (\ref{pro1_ea1}) can be reformulated as follows.
\begin{equation}\label{ea7}
\begin{gathered}
\mathop {\min }\limits_{U} \mathop \sum \limits_{i = 1}^n \mathop \sum \limits_{j = 1}^c {u_{ij}}\bigg\{  m{{ {{\left[ {{{(x_i  - {v_j})}^T}{\Sigma_j^{ - 1}}(x_i - {v_j})} \right]}} }^\beta }
\\ \quad \ \ + \frac{1}{2}\log |{\Sigma _j}| - \log {\alpha _j} + \log {u_{ij}} \bigg\},
\\ \ \ \ \ \ {\rm s.t.} \ \ \mathop \sum \limits_{j = 1}^c {u_{ij}} = 1, \ \ 0 < {u_{ij}} < 1
\end{gathered}
\end{equation}
We remove the inequality constraints and take the Lagrange multiplier method to optimize the objective function in Eq. (\ref{ea7}). The corresponding Lagrange function is constructed as
\begin{equation}\label{ap12}
\begin{gathered}
L(U,\gamma) =  \mathop \sum \limits_{i = 1}^n  \mathop \sum \limits_{j = 1}^c  {u_{ij}}\bigg\{  m{{ {{\left[ {{{(x_i  - {v_j})}^T}{\Sigma_j^{ - 1}}(x_i - {v_j})} \right]}} }^\beta } \\
+ \frac{1}{2}\log |{\Sigma _j}| - \log {\alpha _j} + \log {u_{ij}} \bigg\}  + \mathop \sum \limits_{i = 1}^n \eta_i ( \mathop \sum \limits_{j = 1}^c {u_{ij}} - 1)
\end{gathered}
\end{equation}
where $\eta = \{ \eta_1, \eta_2,..., \eta_n \}$ is the set of Lagrange multipliers. By setting the derivative of the Lagrange function in Eq. (\ref{ap12}) to zero with respect to $u_{ij}$ we have
\begin{equation}\label{ap13}
\begin{aligned}
\frac{\partial L(U,\eta)}{\partial u_{ij}} = m{\left[ {{{(x_i - {v_j})}^T}\Sigma _j^{ - 1}(x_i - {v_j})} \right]^\beta } \\ +  \frac{1}{2}\log |{\Sigma _j}|  - \log {\alpha _j} + \log {u_{ij}} + 1 + {\eta _i} =0
\end{aligned}
\end{equation}
Then according to Eq. (\ref{ap13}) we can obtain
\begin{equation}\label{ap131}
u_{ij} \! = \!  \frac{{{\alpha _j}|{\Sigma _j}{|^{ - \frac{1}{2}}}\exp \left\{  - m{{\left[ {{{({x_i} \!-\! {v_j})}^T}\Sigma _j^{ - 1}({x_i} \!-\! {v_j})} \right]}^\beta }\right\} }}{{\exp ( 1 + {\eta _i}) }}
\end{equation}
By combining Eq. (\ref{ap131}) with the constraint $\sum_{j=1}^{c} u_{ij}=1$, we can obtain
\begin{equation}\label{ap132}
\begin{gathered}
\exp ( 1+\eta_i ) \! =\!\! \mathop \sum \limits_{j = 1}^c \!  {\alpha _j}|{\Sigma _j}|^{\frac{{ - 1}}{2}} \times  \quad \qquad\qquad\qquad          \\
\quad\qquad\qquad \exp \left\{  - m {{\left[ {{{(x_i - {v_j})}^T}\Sigma _j^{ - 1}\!(x_i - {v_j})}  \right]}^\beta }  \right\}
\end{gathered}
\end{equation}
Substitute $\exp\{1+\eta_i\}$ in Eq. (\ref{ap132}) back into Eq. (\ref{ap131}) and then we can obtain
\begin{equation}\label{ap14}
\hspace{-0.18cm}
u_{ij}\!\! = \!\! \frac{{{\alpha _j}|{\Sigma _j}{|^{\frac{{ - 1}}{2}}} \! \exp \left\{ \! - m {{\left[ {{{(x_i \!-\! {v_j})}^T}\Sigma _j^{ - 1}(x_i \!-\! {v_j})} \right]}^\beta } \right\} } } {{\mathop \sum \limits_{j = 1}^c \!\! {\alpha _j}|{\Sigma _j}{|^{\frac{{ - 1}}{2}}} \! \exp \left\{  \! - m{{\left[ {{{(x_i \!-\! {v_j})}^T}\Sigma _j^{ - 1}(x_i \!-\! {v_j})} \right]}^\beta } \right\} }}
\end{equation}
The solution of in Eq. (\ref{ap14}) satisfies the inequality constraint $0<u_{ij}<1$, and hence it is a suitable update equation for $u_{ij}$. By comparing the update equation of $u_{ij}$ in Eq. (\ref{ap14}) and the update equation of $p(j|{x_i},{\Theta ^{t - 1}})$ in Eq. (\ref{ea5}), we can find they have the same form. Therefore, $u_{ij}$ can be seen as an equivalence of $p(j|{x_i},{\Theta ^{t - 1}})$.

{\bf In the second part:} When we fix $U$ to update the other variables, the objective function in Eq. (\ref{pro1_ea1}) can be reformulated as follows.

\begin{equation}\label{part2}
\begin{gathered}
\mathop {\min }\limits_{\alpha , V, \Sigma, \beta ,m} \mathop \sum \limits_{i = 1}^n \mathop \sum \limits_{j = 1}^c {u_{ij}}\bigg\{ m{{ {{\left[ {{{(x_i  - {v_j})}^T}{\Sigma_j^{ - 1}}(x_i - {v_j})} \right]}} }^\beta }
\\ \qquad \quad + \frac{1}{2}\log |{\Sigma _j}| - \log {\alpha _j} - \log \frac{{\beta \Gamma (\frac{d}{2}){m^{\frac{d}{{2\beta }}}}}}{{{\pi ^{\frac{d}{2}}}\Gamma (\frac{d}{{2\beta }})}} \bigg\}, \\
{\rm s.t.}\ \ \mathop \sum \limits_{j = 1}^c {\alpha _j} = 1, \ \ 0 < {\alpha _j} < 1
\end{gathered}
\end{equation}
By comparing the objective function in Eq. (\ref{part2}) and the Q function in Eq. (\ref{ea6}), it is observed that they have the same form and hence their variables share the same update equations as well.

Up to this point, the proof for the equivalence between the objective function in Eq. (\ref{pro1_ea1}) and the log-likelihood function in Eq. (\ref{ea4}) has been completed.

\section*{proof for Eq. (\ref{emm})} \label{Eq11_proof}
To facilitate the readers, we first rewrite Eq. (\ref{emm}) as below.
\begin{equation}\label{emm_a1}
\begin{gathered}
\sum_{i=1}^n\sum_{j=1}^cu_{ij}||x_i-v_j||_2^2 \mathop  = Tr\left[ X(I_n-UBU^T)X^T \right]
\end{gathered}
\end{equation}
where $X=[x_1,x_2,...,x_n]\in\mathbb{R}^{d\times n}$ denotes the data matrix, $u_{ij}$ denotes the membership degree satisfying $\sum_{j=1}^{c}u_{ij} = 1$, and $B\in\mathbb{R}^{c\times c}$ is a diagonal matrix with the $k$-th diagonal element $b_{kk}= 1/\sum_{i=1}^{n}u_{ik}$, and $v \in \mathbb{R}^d$ denotes the $j$-th cluster center. Moreover, the update equation for $v_j$ is supposed to be
\begin{equation}\label{v_up}
v_j= \frac{\sum_{i=1}^{n}u_{ij}x_i}{\sum_{i=1}^{n}u_{ij}} =
\sum_{i=1}^n{\left( \frac{u_{ij}}{\sum_{l=1}^n{u_{lj}}} \right) x_i}
\end{equation}
We define
\begin{equation}\label{p_dfine}
\begin{gathered}
p_{ij}=\frac{u_{ij}}{\left( \sum_{l=1}^n{u_{lj}} \right)}, \ \ P_j=\left( \begin{matrix}
	p_{1j}&		&		0\\
	&		\ddots&		\\
	0&		&		p_{nj}\\
\end{matrix} \right) \in \mathbb{R} ^{n\times n}
\end{gathered}
\end{equation}
where $P_j$ is a diagonal matrix with the $k$-th diagonal element being $p_{kj}$.
Then we have
\begin{equation}\label{v_trans}
v_j = \sum_{i=1}^{n}p_{ij}x_i
\end{equation}
Let us now direct the attention to the left-hand side of Eq. (\ref{emm_a1}).
\begin{equation}\label{emm_a2}
\begin{aligned}
&\sum_{i=1}^n\sum_{j=1}^cu_{ij}||x_i-v_j||_2^2 \\
=&Tr\left[ \sum_{i=1}^n{\sum_{j=1}^c{u_{ij}(x_i-v_j)(x_i-v_j)^T}} \right] \\
=&Tr\left[ \sum_{j=1}^c{\left( \sum_{l=1}^n{u_{lj}} \right) \sum_{i=1}^n{\frac{u_{ij}}{\left( \sum_{l=1}^n{u_{lj}} \right)}(x_i-v_j)(x_i-v_j)^T}} \right]\\
=&Tr\left[ \sum_{j=1}^c{\left( \sum_{l=1}^n{u_{lj}} \right) \sum_{i=1}^n{p_{ij}(x_i-v_j)(x_i-v_j)^T}} \right]
\end{aligned}
\end{equation}
According to the definition of $p_{ij}$ and $P_j$ in Eq. (\ref{p_dfine}) and the definition of $v_j$ in Eq. (\ref{v_trans}), we can further conclude
\begin{equation}\label{m_trans}
\begin{aligned}
	&\sum_{i=1}^n{p_{ij}(x_i-v_j)(x_i-v_j)^T}\\
	=&\left[ X\left( I_n-P_j{\bf1}{\bf1}^T \right) \right] P_j\left[ X\left( I_n-P_j{\bf1}{\bf1}^T \right) \right] ^T\\
	=&X\left( I_n-P_j{\bf1}{\bf1}^T \right) P_j\left( I_n-{\bf1}{\bf1}^TP_j \right) X^T\\
	=&X\left( P_j-P_j{\bf1}{\bf1}^TP_j \right) \left( I_n-{\bf1}{\bf1}^TP_j \right) X^T\\
	=&X\left( P_j-P_j{\bf1}{\bf1}^TP_j-P_j{\bf1}{\bf1}^TP_j+P_j{\bf1}{\bf1}^TP_j{\bf1}{\bf1}^TP_j \right) X^T\\
	=&X\left( P_j-P_j{\bf1}{\bf1}^TP_j \right) X^T\\
\end{aligned}
\end{equation}
where ${\bf1}\in\mathbb{R}^{n\times 1}$ is a vector with all elements being equal to $1$. Substitute Eq. (\ref{m_trans}) back into Eq. (\ref{emm_a2}) and then we have
\begin{equation}\label{final}
\begin{aligned}
&Tr\left[ \sum_{j=1}^c{\left( \sum_{l=1}^n{u_{lj}} \right) \sum_{i=1}^n{p_{ij}(x_i-v_j)(x_i-v_j)^T}} \right]
\\
=&Tr\left[ \sum_{j=1}^c{\left( \sum_{l=1}^n{u_{lj}} \right) X(P_j-P_j11^TP_j)X^T} \right]
\\
=&Tr\left\{ X\left[ \sum_{j=1}^c{\left( \sum_{l=1}^n{u_{lj}} \right) P_j-\sum_{j=1}^c{\left( \sum_{l=1}^n{u_{lj}} \right)}P_j11^TP_j} \right] X^T \right\}
\\
=&Tr\left[ X\left( I_n-UBU^T \right) X^T \right]
\end{aligned}
\end{equation}
where $B\in\mathbb{R}^{c\times c}$ is a diagonal matrix with the definition given as below.
\begin{equation}
B=\left( \begin{matrix}
	\frac{1}{\sum_{l=1}^n{u_{l1}^{}}}&		&		0\\
	&		\ddots&		\\
	0&		&		\frac{1}{\sum_{l=1}^n{u_{lc}^{}}}\\
\end{matrix} \right)
\end{equation}
Up to this point, the proof for Eq. (\ref{emm}) has been completed.

}

\bibliographystyle{IEEEtran}
\bibliography{reference}
%

\newpage

\vfill

\end{document}